\documentclass{article} % For LaTeX2e
\usepackage{iclr2024_conference,times}
\iclrfinalcopy

\usepackage{xspace}
\usepackage{times}
\usepackage{epsfig}
\usepackage{graphicx}
\usepackage{amsmath}
\usepackage{amssymb}
\usepackage{tabularx}

\usepackage[utf8]{inputenc}
\usepackage{multirow}
\usepackage{rotating}
\usepackage{verbatim}
\usepackage{booktabs}
\usepackage{array}
\usepackage{textcomp, gensymb}
\usepackage{diagbox}
\usepackage[T1]{fontenc}    % use 8-bit T1 fonts
\usepackage{url}            % simple URL typesetting
\usepackage{booktabs}       % professional-quality tables
\usepackage{amsfonts}       % blackboard math symbols
\usepackage{nicefrac}       % compact symbols for 1/2, etc.
\usepackage{microtype}      % microtypography
\usepackage{setspace}
\usepackage{multirow}
\usepackage{amssymb}
\usepackage{diagbox}
\usepackage{mathtools}
\usepackage{wrapfig}
\usepackage{diagbox} 
\usepackage{wrapfig,lipsum,booktabs}
\usepackage{caption}
\usepackage{appendix}
\usepackage{tikz}

\usepackage{colortbl}
\usepackage{xcolor}
\usepackage{color}
\definecolor{citecolor}{RGB}{0,113,188}
\definecolor{table_second}{HTML}{FFEDB3}
\definecolor{table_best}{HTML}{FFD44D}
\definecolor{yellow_1}{HTML}{EEB422}
\definecolor{yellow_2}{HTML}{FFD700}

\usepackage{soul}
\DeclareRobustCommand{\hlYellowOne}[1]{{\sethlcolor{table_best}\hl{#1}}}
\DeclareRobustCommand{\hlYellowTwo}[1]{{\sethlcolor{table_second}\hl{#1}}}
\colorlet{soulblue}{citecolor!10}
\DeclareRobustCommand{\hlBlue}[1]{{\sethlcolor{soulblue}\hl{#1}}}

\usepackage{colortbl}
\usepackage{array}

\newcolumntype{x}[1]{>{\centering\arraybackslash}p{#1pt}}

\newcommand{\app}{\raise.17ex\hbox{$\scriptstyle\sim$}}

\newlength\savewidth\newcommand\shline{\noalign{\global\savewidth\arrayrulewidth
  \global\arrayrulewidth 1pt}\hline\noalign{\global\arrayrulewidth\savewidth}}
\newcommand{\tablestyle}[2]{\setlength{\tabcolsep}{#1}\renewcommand{\arraystretch}{#2}\centering\footnotesize}

\newcommand{\tablefirst}{\cellcolor{table_best}}
\newcommand{\tablesecond}{\cellcolor{table_second}}
\newcommand{\tablegtfirst}{\cellcolor{citecolor!10}}

\usepackage{amssymb}% http://ctan.org/pkg/amssymb
\usepackage{pifont}% http://ctan.org/pkg/pifont

\newcommand{\cmark}{\ding{51}}%
\newcommand{\xmark}{\ding{55}}%

\newcommand{\modelname}[0]{\mbox{\textsc{LEAP}}\xspace}

\usepackage{amsmath,amsfonts,bm}

\def\eqref#1{equation~\ref{#1}}
\def\1{\bm{1}}

\DeclareMathAlphabet{\mathsfit}{\encodingdefault}{\sfdefault}{m}{sl}
\SetMathAlphabet{\mathsfit}{bold}{\encodingdefault}{\sfdefault}{bx}{n}

\usepackage[pagebackref=false,breaklinks=true,letterpaper=true,citecolor=citecolor,colorlinks,bookmarks=false]{hyperref}
\usepackage{url}
\usepackage{graphicx}

\title{\modelname{}: \underline{L}iberate Spars\underline{e}-view 3D Modeling \\from Camer\underline{a} \underline{P}oses}

\author{Hanwen Jiang \quad Zhenyu Jiang \quad Yue Zhao \quad Qixing Huang \\
  Department of Computer Science,
  The University of Texas at Austin\\
  \small{\hspace{0mm}Project page: \href{https://hwjiang1510.github.io/LEAP/}{https://hwjiang1510.github.io/LEAP/}}
}

\begin{document}

\maketitle

\begin{figure}[h]
    \centering
    \vspace{-0.2in}
    \includegraphics[width=\linewidth]{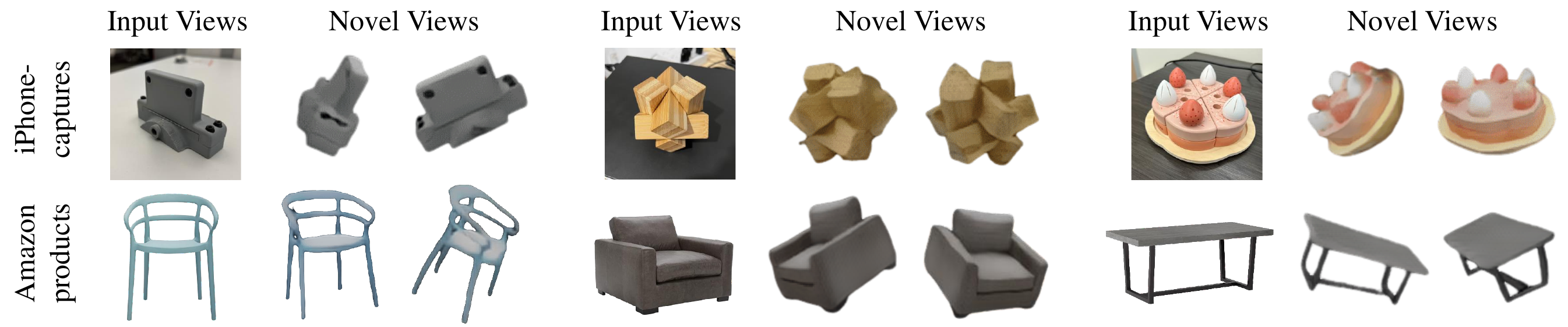}
    \vspace{-0.27in} \caption{\small{\modelname{} performs 3D modeling from \textbf{sparse views without camera pose information}. We show the capability of \modelname{} on real-world cases with \textbf{three unposed image inputs}. We show one of the inputs.}}
\end{figure}

\begin{abstract}
Are camera poses necessary for multi-view 3D modeling?
Existing approaches predominantly assume access to accurate camera poses. While this assumption might hold for dense views, accurately estimating camera poses for sparse views is often elusive.
Our analysis reveals that noisy estimated poses lead to degraded performance for existing sparse-view 3D modeling methods.
To address this issue, we present \modelname{}, a novel \textit{pose-free} approach, therefore challenging the prevailing notion that camera poses are indispensable.
\modelname{} discards pose-based operations and learns geometric knowledge from data. \modelname{} is equipped with a neural volume, which is shared across scenes and is parameterized to encode geometry and texture priors. For each incoming scene, we update the neural volume by aggregating 2D image features in a feature-similarity-driven manner. The updated neural volume is decoded into the radiance field, enabling novel view synthesis from any viewpoint.
On both object-centric and scene-level datasets, we show that \modelname{} significantly outperforms prior methods when they employ predicted poses from state-of-the-art pose estimators.
Notably, \modelname{} performs on par with prior approaches that use ground-truth poses while running $400\times$ faster than PixelNeRF.
We show \modelname{} generalizes to novel object categories and scenes, and learns knowledge closely resembles epipolar geometry.
\end{abstract}

% Existing 3D modeling approaches predominantly assume access to accurate camera poses. While this assumption might hold for dense views, accurately estimating camera poses for sparse views is often elusive. Our analysis reveals that noisy estimated poses lead to degenerated performance for existing sparse-view 3D modeling methods.
% To address this issue, we present \modelname{}, a novel \textit{pose-free} approach to 3D modeling.
% \modelname{} discards pose-based operations and learns geometric knowledge from data. \modelname{} leverages attention to aggregate information from 2D images and map it into a 3D feature grid with a neural volume shared across scenes. The 3D feature grid is decoded into a radiance field, enabling novel view synthesis from any viewpoint.
% On both object-centric and scene-level datasets, we show that \modelname{} significantly outperforms prior methods when they employ predicted poses from state-of-the-art pose estimators.
% Notably, \modelname{} performs on par with prior approaches that use ground-truth poses while running $400\times$ faster than PixelNeRF.
% In addition, \modelname{} can generalize to novel object categories and scenes. Our in-depth analysis demonstrates that \modelname{} learns geometric prior closely related to epipolar geometry.

\vspace{-2mm}
\section{Introduction}
\vspace{-2mm}

In 3D vision, camera poses offer powerful explicit geometric priors to connect 3D points and 2D pixels~\citep{Zisserman2001MultipleVG}. Its effectiveness has been verified across a spectrum of 3D vision tasks~\citep{goesele2006multi, Geiger2011StereoScanD3}, enabling high-quality 3D modeling~\citep{Mildenhall2020NeRFRS, Wang2021NeuSLN}.
%Prior 3D modeling works predominantly assume access to highly accurate camera poses for testing their capability~\citep{Mildenhall2020NeRFRS, pixelnerf}.
However, accurate camera poses are not always available in the real world, and inaccurate poses lead to degraded performance~\citep{Lin2021BARFBN}. To obtain accurate camera poses, one solution is capturing \textit{dense} views and applying structure-from-motion techniques~\citep{schoenberger2016sfm}.
Nevertheless, in real-world scenarios, like product images in online stores, we usually observe \textit{sparse} images captured by wide-baseline cameras. For sparse views, estimating accurate camera poses is still challenging~\citep{Zhang2022RelPose}.
Then a question arises: is using noisy estimated camera poses still the best choice for \textbf{3D modeling from sparse and unposed views}?

In this paper, we present \modelname{}, which champions a \textit{pose-free} paradigm. Instead of pursuing a more accurate camera pose estimator, \modelname{} challenges the prevailing notion that camera poses are indispensable for 3D modeling. \modelname{} abandons any operations that explicitly use camera poses, e.g. projection, and learns the pose-related geometric knowledge/representations from data.
Thus, \modelname{} is entirely liberated from camera pose errors, leading to better performance.

\modelname{} specifically represents each scene as a neural radiance field, which is predicted in a single feed-forward step. To initialize the radiance field, we introduce a neural volume, which is shared across all scenes. Each voxel grid of the volume is parameterized to learn the geometry and texture priors from data. For any incoming scene, the neural volume queries input 2D image features and gets updated through aggregation.
Instead of using camera poses to identify source 2D pixels to aggregate ~\citep{pixelnerf}, \modelname{} leverages attention to aggregate all 2D image features with adaptive weights.
Subsequently, \modelname{} performs spatial-aware attention on the updated neural volume to capture long-range geometry dependency. We iterate the process of aggregating and 3D reasoning, resulting in a refined neural volume. The refined neural volume is then decoded into the radiance field.

An important issue is which reference 3D coordinate frame we should use to define the neural volume. A good choice of this 3D coordinate frame can significantly stabilize and enhance learning~\citep{Qi2016PointNet, Deng2021VectorNerons}. 
As the world coordinate frame for an unposed image set is not well-defined, we instead use a local coordinate frame. Specifically, we choose an arbitrary input image as the canonical view, and the neural volume is defined in its corresponding local camera coordinate. The camera pose of the canonical view is fixed, e.g. as an identity pose, in the local camera coordinate frame. To enable the model aware of the choice of canonical view, we find the key is making 2D image features of non-canonical views consistent with the canonical view. Thus, we design a multi-view encoder to improve the consistency by capturing cross-view 2D correlations.

During training, the canonical view is randomized among all input views. We train \modelname{} with 2D rendering loss of the input views, using ground-truth camera poses to render them. Note that these ground-truth camera poses are only used during training to learn the mapping from input images to the neural volume. During inference, \modelname{} predicts the radiance field without reliance on any poses.

We perform a thorough evaluation of \modelname{} on a diverse array of object-centric~\citep{Wu2023OmniObject3DL3, forge, Deitke2022ObjaverseAU} and scene-level~\citep{Jensen2014LargeSM} datasets. This assessment spans multiple data scales and incorporates both synthetic and real images.
Experimental results highlight \modelname{}'s four interesting properties: i) \textbf{Superior performance}. \modelname{} consistently synthesizes novel views from $\mathbf{2\sim5}$ unposed images. It surpasses prior generalizable NeRFs when they use camera poses predicted by SOTA pose estimators. It performs on par with methods using ground-truth camera poses. ii) \textbf{Fast inference speed}. \modelname{} constructs the radiance field in a feed-forward manner without optimization, running within one second on a single consumer-grade GPU. iii) \textbf{Strong generalization capability}. \modelname{} models novel-category objects accurately. The model learned on large object-centric datasets transfer well to the scene-level DTU dataset. iv) \textbf{Interpretable learned priors}. While \modelname{} does not explicitly use camera poses by design, it acquires priors consistent with epipolar geometry.
We are committed to \textit{releasing code} for reproducibility and future research.

\vspace{-2mm}
\section{Related Work}
\vspace{-2mm}

\paragraph{NeRF from sparse views with ground-truth camera poses.} NeRF variants that work on sparse view inputs can be categorized into two genres. The first is \textbf{scene-specific NeRFs}. Following the original NeRF setting, these methods optimize the radiance field for each scene from scratch. They use additional information to regularize NeRF optimization, e.g., normalization flow~\citep{Niemeyer2021RegNeRF} and frequency regularization~\citep{Yang2023FreeNeRF}. The second is \textbf{generalizable NeRF variants}~\citep{pixelnerf, Wang2021IBRNet, Chen2021MVSNeRF}, which predict the radiance field in a feed-forward manner. The key is making the radiance fields conditioned on the 2D image features. Typically, these approaches project the 3D points on the input images using camera poses, and information is aggregated from the image features at projected 2D locations. Thus, they are generalizable and transferable to novel scenes by training on curated datasets. However, these methods lack reasoning of correlations between 3D points and assume the access of ground-truth camera poses. In contrast, \modelname{} has 3D reasoning ability, which works on images without poses.

\vspace{-0.15in}
\paragraph{Sparse-view camera pose estimation.}
Estimating the camera poses from sparse views 
presents a significantly greater challenge than from dense views. The complexity arises from the minimal or absent overlap between images, which hampers the formation of cross-view correspondence cues, vital for accurate camera pose estimation~\citep{Zisserman2001MultipleVG}. RelPose~\citep{Zhang2022RelPose} highlights the limitations of conventional dense-based camera pose estimation techniques, e.g., COLMAP~\citep{schoenberger2016sfm}, in sparse view contexts. In response, it introduces an energy-based model to handle multi-modal solutions indicative of potential camera poses. A subsequent method~\citep{relpose2} further develops this approach by harnessing multi-view information to enhance pose estimation accuracy. Concurrently, SparsePose employs a pre-trained foundational model, namely DINO~\citep{Caron2021EmergingPI}, to iteratively refine the predictions of noisy camera poses. 
Besides, researchers also have explored  directional representation~\citep{chen2021widebaseline1}, stronger image matching techniques~\citep{sun2021loftr} or using image matching priors~\citep{rockwell202288point}, and co-visibility~\citep{hutchcroft2022covispose} to improve the performance.
In contrast, our \modelname{} operates without dedicated camera pose estimation modules.

%SAMURAI~\citep{boss2022samurai} proposes to refine noisy poses by jointly optimizing them with the geometry but requires dense views for training on each object instance from scratch.

\vspace{-0.15in}
\paragraph{NeRF with imperfect or no camera poses.}
Building NeRF from images without precise camera poses is challenging, given that many NeRF variants rely on pose-based geometric operations. To tackle this problem, scene-specific NeRFs~\citep{Lin2021BARFBN, Wang2021NeRF--, Xia2022SiNeRFSN} treat camera poses as modifiable parameters, jointly optimizing them alongside the radiance field. Yet, these methods require dense views, and they either require reasonably accurate initial poses or assume small-baseline cameras to work. 
NeRS~\citep{zhang2021ners} employs surface-based 3D representations as optimization regularizers, suitable for sparse views. However, it is restricted to object-centric data and requires category-specific shape templates. SPARF~\citep{Truong2022SPARFNR} leverages dense 2D image correspondences derived from existing models to augment radiance field optimization. Nevertheless, its efficacy heavily hinges on the precision of both dense correspondences and initial poses.
For generalizable NeRF variants, SRT~\citep{srt} proposes a pose-free paradigm building a 2D latent scene representation. However, SRT is not 3D-aware, and its novel view synthesis quality is limited. FORGE~\citep{forge} jointly estimates camera poses and predicts the radiance field, leveraging their synergy to improve the performance of both.
However, the performance is sensitive to pose estimation precision and training FORGE in multi-stages is non-trivial.
In contrast, our proposed \modelname{} benefits from the 3D-aware designs and leans into a feature-similarity-driven 2D-3D information mapping. This approach eliminates reliance on camera poses, yielding results more closely aligned with using ground-truth poses.

\begin{figure*}[t]
\centering
\vspace{-0.2in}
\includegraphics[width=0.95\linewidth]{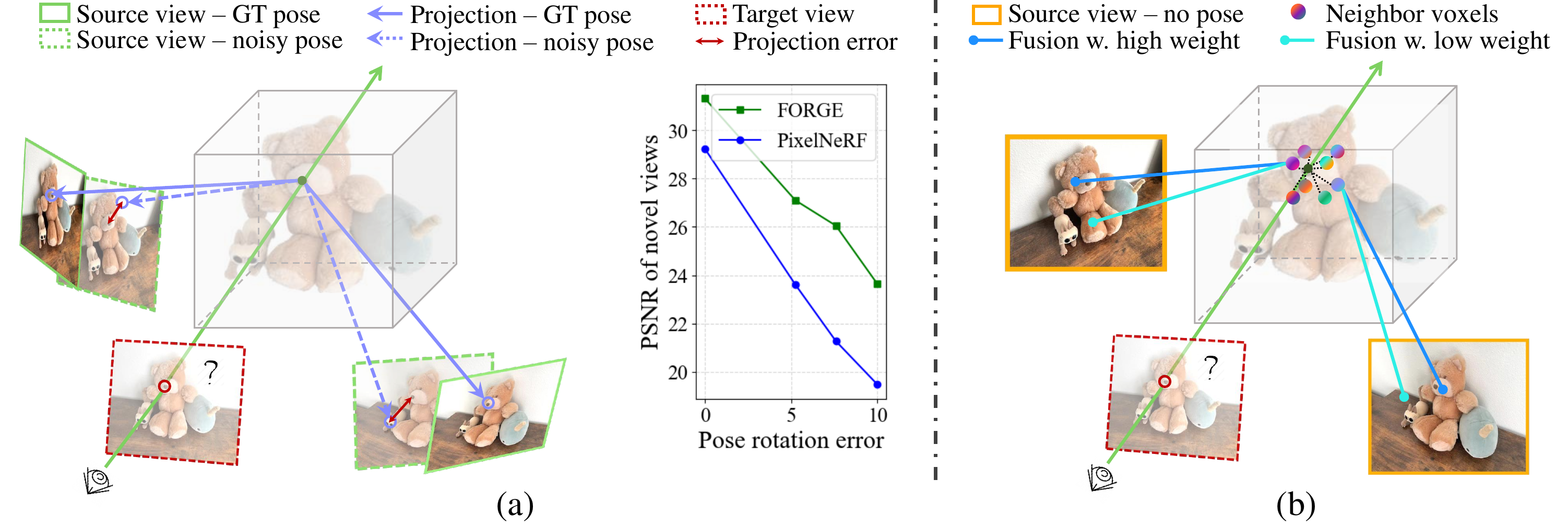}
\vspace{-0.1in}
\caption{\small{
\textbf{Comparing \modelname{} with pose-based generalizable NeRFs~\citep{pixelnerf}.} (a) For any 3D point along the casting ray from a target view, pose-based methods project these 3D points onto input source views and aggregate 2D image features at these projected locations.
These methods are sensitive to slight pose inaccuracies, as the errors cause misaligned 3D points and 2D pixels.
(b) In contrast, \modelname{} offers a pose-free method. It defines a parametrized 3D neural volume to encapsulate geometry and texture priors. For any scene, each voxel grid aggregates information from all 2D image pixel, and their feature similarity determines the fusion weight. Features for an inquired 3D point are interpolated from the neighboring voxel grids. This sidestep of pose-dependent operations allows for direct inference on unposed images.
}}
\vspace{-0.15in}
\label{fig: overview}
\end{figure*}

\vspace{-2mm}
\section{Overview}
\vspace{-2mm}

%\yzhao{The first paragraph resemble the related work. Try merge the first two paragraphs into one.}

We focus on novel view synthesis from \textbf{sparse views} of a scene \textbf{without camera poses}. Prior approaches have adjusted NeRF for sparse views under the assumption of accurate camera poses~\citep{pixelnerf, Chen2021MVSNeRF, Niemeyer2021RegNeRF, Yang2023FreeNeRF}. Concurrently, enhanced camera pose estimation methods for these sparse images have emerged~\citep{Zhang2022RelPose}. However, preliminary results for combining the efforts indicate a potential incompatibility; minor pose estimation inaccuracies can significantly degrade the quality of synthesized views in NeRF~\citep{Truong2022SPARFNR, forge, sparsepose}.

We first diagnose the limitations of existing approaches. As illustrated in Fig.~\ref{fig: overview}, the existing generalizable NeRF approaches~\citep{pixelnerf, Wang2021IBRNet} rely on camera poses for performing 2D-3D information mapping.  Specifically, these methods \textit{project a 3D point} to its single corresponding 2D locations in each of the input images \textit{based on camera poses}, and aggregate features at these projected locations. Consequently, any pose inaccuracies distort this 3D-2D association, leading to compromised 3D point features, which are used to predict the radiance.

In contrast, \modelname{} proposes a novel \textit{pose-free} paradigm, eliminating the influence of any pose inaccuracies. At its core, \modelname{} establishes the 3D-2D association based on \textit{feature similarities}, enabling a 3D point to aggregate information from all pixels, rather than its 2D projection only. For each 3D voxel grid, the pose-free aggregation will learn to adaptively assign larger weights to its corresponding 2D pixels. We introduce the details of \modelname{} architecture in the following section.

\begin{figure*}[t]
\centering
\includegraphics[width=\linewidth]{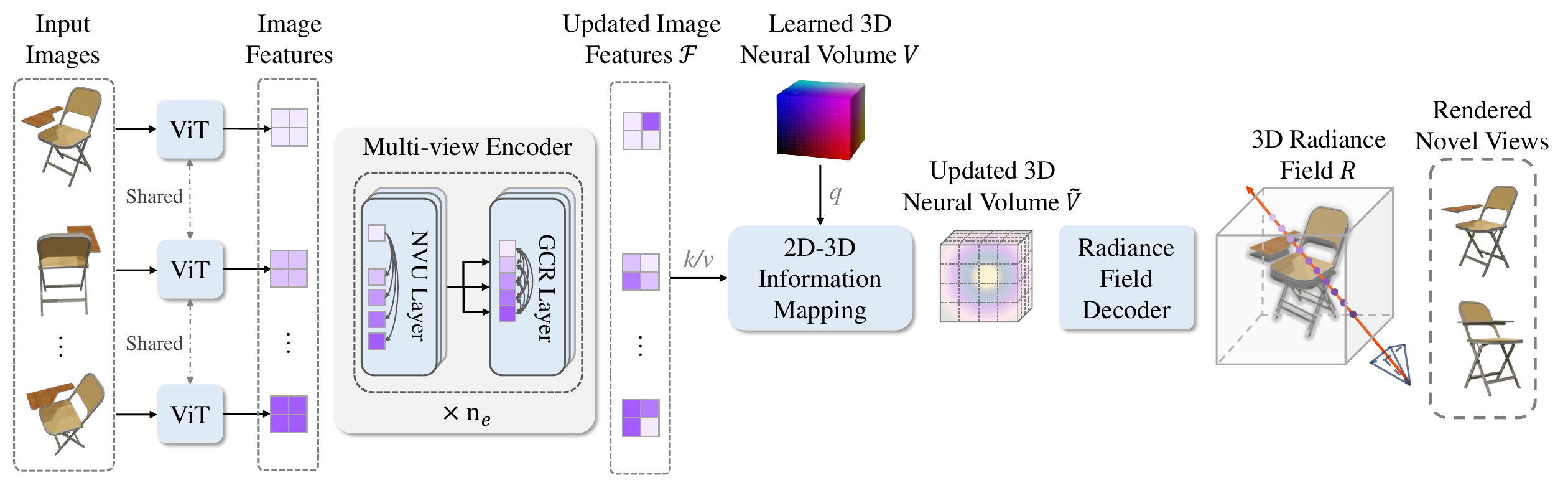}
\vspace{-0.25in}
\caption{\small{\textbf{\modelname{} overview.} \modelname{} extracts image features of all inputs using a ViT backbone. The first image in the image set is selected as the canonical view. The neural volume is defined in the local camera coordinate of the canonical view, which has learnable parameters to encode geometry and texture priors. To enable \modelname{} aware of the choice of canonical view, we use a Multi-View Encoder to propagate information from the canonical view to the non-canonical views, making the 2D representations more coherent across views. Then the neural volume is updated by querying the 2D image features of all images using a 2D-3D Information Mapping module. We decode the neural volume into the radiance field and render novel views at inference time. %\yzhao{shorter and more concise}
 }}
\vspace{-0.12in}
\label{fig: main}
\end{figure*}

\vspace{-2mm}
\section{Method}
\vspace{-2mm}

We present the task formulation and the overview of \modelname{}. Given a set of $k$ 2D image observations of a scene, represented as $\{I_i \vert i=1,...,k\}$, \modelname{} predicts a neural radiance field~\citep{Mildenhall2020NeRFRS}, which enables synthesizing a 2D image from an arbitrary target viewpoint. 
Note that in our setup of \textit{sparse source views} captured by wide-baseline cameras, the number $k$ is typically less than $5$. 
Moreover, these views are presented \textit{without any associated camera pose information} at inference.

\subsection{Model Architecture}

As illustrated in Fig.~\ref{fig: main}, \modelname{} initiates by extracting 2D image features from all views. 
We use a DINOv2-initialized ViT~\citep{Oquab2023DINOv2LR, vit} as the feature extractor since it demonstrates strong capability in modeling cross-view correlations~\citep{Zhang2023Ataleoftwofeat}. 
We denote the image features of $I_i$ as $f_i \in \mathbb{R}^{h\times w\times c}$, and
the resulting features set for all input views as $\{f_i \vert i=1,...,k\}$.
Due to the unawareness of the world coordinate frame on unposed images, we perform 3D modeling in a local camera coordinate frame. Specifically, we designate one image as the canonical view,
where the neural volume and radiance field are defined in its local coordinate frame. During training, we randomly select a canonical view and denote it as $I_1$ for notation clarity. 

To make \modelname{} aware of the choice of the canonical view, we find the key is to make the features of the non-canonical views consistent with the canonical view.
We propose a multi-view image encoder to improve the feature consistency.
Then, \modelname{} introduces a learnable neural volume, which is shared across scenes, to encode the geometric and textural prior. The neural volume serves as the initial 3D representation for all scenes. 
For each incoming scene, by querying the multi-view features, \modelname{} maps the 2D information to the 3D domain, represented by an updated neural volume. 
Finally, \modelname{} predicts the radiance field from the updated neural volume.
We describe each step as follows.
%For training, \modelname{} utilizes a photometric loss that measures the disparity between the input images and their rendered counterparts (Sec.~\ref{sec: train}).

%\vspace{-0.15in}
\noindent\textbf{Multi-view Image Encoder} enables \modelname{} aware of the choice of the canonical view, by performing multi-view information reasoning.
It takes in image features of all views and refines them by capturing cross-view correlations.
%which make \modelname{} aware of the choice of canonical view. 
It consists of $n_e$ blocks, and each block has two layers: a Non-canonical View Update (NVU) layer and a Global Consensus Reasoning (GCR) layer. The NVU layer updates each of the non-canonical view features by aggregating the canonical view features.
It is denoted as $\bar{f}_j = \textrm{NVU}(f_j, f_1)$, where $j\ne 1$ and $\bar{f}$ denotes the updated features.
The GCR layer performs joint reasoning on all views for a global consensus, which leverages the correlation between all views.

We implemented the two layers with Transformer layers~\citep{transformer}.
% , leveraging its strong correlation modeling capability.
Specifically, the NVU layer is modeled as a Transformer layer with cross-attention, where the non-canonical view features are queries, and the canonical view features are keys/values. 
It is formulated as
\begin{align}
    \left[\bar{f}_2, ..., \bar{f}_k\right] = \text{FFN}(\text{CrossAttention}(\left[f_2, ..., f_k\right], f_1)),
\end{align}
where FFN is a feed-forward network, and $\left[ \cdot \right]$ denotes the concatenating operation over tokenized image features. For clarity, $\left[\bar{f}_2, ..., \bar{f}_k\right]$ is in $\mathbb{R}^{(k-1)hw\times c}$ and $f_1$ is flattened into $\mathbb{R}^{hw\times c}$.

Similarly, the GCR layer is instantiated by a Transformer layer with self-attention, where the query, key, and value are the 2D image features of all views. It is formulated as 
\begin{align}
    \left[\tilde{f}_1, ..., \tilde{f}_k\right] = \text{FFN}(\text{SelfAttention}(\left[f_1, \bar{f}_2, ..., \bar{f}_k\right])).
\end{align}
Specifically, $\left[\tilde{f}_1, ..., \tilde{f}_k\right]$ is in $\mathbb{R}^{khw\times c}$. 
For simplicity, we denote the final output $\left[\tilde{f}_1, ..., \tilde{f}_k\right]$ as $\mathbf{F}$.

\noindent\textbf{2D-3D Information Mapping.}
\modelname{} introduces a 3D latent neural volume $ \mathbf{V} \in \mathbb{R}^{H\times W\times D\times c} $ to encode the geometry and texture priors, where $H$, $W$, $D$ are the spatial resolution of the volume.
It is defined in the local camera coordinate of the canonical view. 
The neural volume is shared across different scenes and gets updated by mapping the 2D image information to the 3D domain.
%The physical size and the origin of the neural volume are hyperparameters.

To perform the 2D-3D information mapping, we use $n_m$ Transformer Decoder blocks~\citep{transformer}, each of which consists of a cross-attention layer and a self-attention layer.
In the cross-attention layer, we use the 3D latent volume $\mathbf{V}$ as the query and use $\mathbf{F}$ as the key/value. The updated 3D neural volume $\bar{\mathbf{V}}$ is defined as $\bar{\mathbf{V}} = \text{FFN}(\text{CrossAttention}(\mathbf{V}, \mathbf{F}))$.
% \begin{align}
%     \bar{v} = \text{FFN}(\text{CrossAttention}(v, \mathcal{F}^{2D})),\quad \bar{v} \in \mathbb{R}^{H\times W\times D\times c}, 
% \end{align}
Intuitively, for each 3D point belonging to the neural volume, we compute its feature similarity with all 2D image features and use the similarity to get the weighted average of 2D image features.
Subsequently, the self-attention layer performs refinement on the 3D volume features, capturing long-range geometry correlation. With multiple 2D-3D information lifting blocks, we reconstruct the latent volume of the objects in a coarse-to-fine manner. We denote the updated neural volume as $\mathbf{\tilde{V}} \in \mathbb{R}^{H\times W\times D\times c}$.

\noindent\textbf{Neural Rendering.}
With the updated neural volume $\mathbf{\tilde{V}}$, \modelname{} predicts a volume-based neural radiance field~\citep{Yu2021PlenoxelsRF, forge}. The radiance field is denoted as 
$\mathbf{R} := (\mathbf{R}_\sigma, \mathbf{R}_f)$, where $\mathbf{R}_\sigma$ and $\mathbf{R}_f$ are the density and features of the radiance field. $\mathbf{R}_\sigma \in \mathbb{R}^{H'\times W'\times D'}$ and $\mathbf{R}_f \in \mathbb{R}^{H'\times W'\times D'\times C}$, where
$H'$, $W'$, and $D'$ are spatial resolutions.
Both $\mathbf{R}_\sigma$ and $\mathbf{R}_f$ are predicted from $\mathbf{\tilde{V}}$ using 3D convolution layers.
We read out the rendered image $\hat{I}$ and object mask $\hat{\sigma}$ using volumetric rendering techniques~\citep{Mildenhall2020NeRFRS}. In detail, we first render a feature map and predict the rendered image using 2D convolutions.
Formally, $(\hat{I}, \hat{\sigma}) = \Pi(\mathbf{R}, \Phi)$, where $\Pi$ denotes the volumetric rendering process, and $\Phi$ is the target camera pose.

\vspace{-2mm}
\subsection{Training and Inference of \modelname{}}
\vspace{-2mm}
\label{sec: train}

\noindent\textbf{Training.} \modelname{} is trained with the photo-metric loss between the rendering results and the inputs without any 3D supervision. We first define the loss $L_{I}$ applied on the RGB images, where $L_{I} = \Sigma_i L_{mse}(\hat{I}_i, I_i) + \Sigma_i \lambda_p L_{p}(\hat{I}_i, I_i)$. The $L_{mse}$ is the MSE loss, $I_i$ and $\hat{I}_i$ are the original and rendered input images, $\lambda_p$ is a hyper-parameter used for balancing losses, and $L_p$ is the perceptual loss~\citep{perceptualLoss}. We then define the loss $L_{M}$ applied on the density masks, as $L_{M} = \Sigma_i L_{mse}(\hat{\sigma}_i, \sigma_i)$, where $\hat{\sigma}_i$ and $\sigma_i$ are original and rendered density masks. The final loss is defined as $L = L_{I} + \lambda_{m} L_{M}$, where $\lambda_{m}$ is the weight balancing hyperparameter. We only use $L_{I}$ if the masks are not available.

\noindent\textbf{Inference and Evaluation.}
During inference, \modelname{} predicts the radiance field without reliance on any poses. To evaluate the novel view synthesis quality, we use the testing camera poses to render the radiance field under specific viewpoints.

% \begin{table*}[t]
%     \tiny
%     \centering
%     %\tablestyle{2pt}{1.1}
%     \fontsize{5pt}{6pt}\selectfont
%     \vspace{-0.05in}
%     %\begin{minipage}[t]{0.95\linewidth}
%     \setlength\tabcolsep{3pt}
%     \begin{tabular}{l|ccc|ccc|ccc|ccc|ccc}
%     & \multicolumn{3}{c|}{\textit{GT Pose}} & \multicolumn{3}{c|}{\textit{Pred. Pose}} & \multicolumn{9}{c}{\textit{No Pose}} \\
 
%     & \multicolumn{3}{c|}{PixelNeRF} 
%     & \multicolumn{3}{c|}{FORGE}
%     & \multicolumn{3}{c|}{Zero-1-to-3}
%     & \multicolumn{3}{c|}{SRT}
%     & \multicolumn{3}{c}{\modelname{}}\\
    
%      &
%     PSNR $\uparrow$ & SSIM $\uparrow$ & LPIPS $\downarrow$ &
%     PSNR $\uparrow$ & SSIM $\uparrow$ & LPIPS $\downarrow$ &
%     PSNR $\uparrow$ & SSIM $\uparrow$ & LPIPS $\downarrow$ &
%     PSNR $\uparrow$ & SSIM $\uparrow$ & LPIPS $\downarrow$ & 
%     PSNR $\uparrow$ & SSIM $\uparrow$ & LPIPS $\downarrow$
%      \\ \shline

%      \textit{Omniobject3D} & 26.97 & 0.888 & 0.123 & 26.56 & 0.889 & 0.109
%     \end{tabular}
%     %\vspace{-0.1in}
%     \caption{\small{Evaluation of novel view synthesis quality on Omniobject3D dataset.}}
%     \label{tabel: omniobject3d}
%     %\end{minipage}
% %\vspace{-0.15in}
% \end{table*}

\begin{figure*}[t]
\centering
\vspace{-0.1in}
\includegraphics[width=\linewidth]{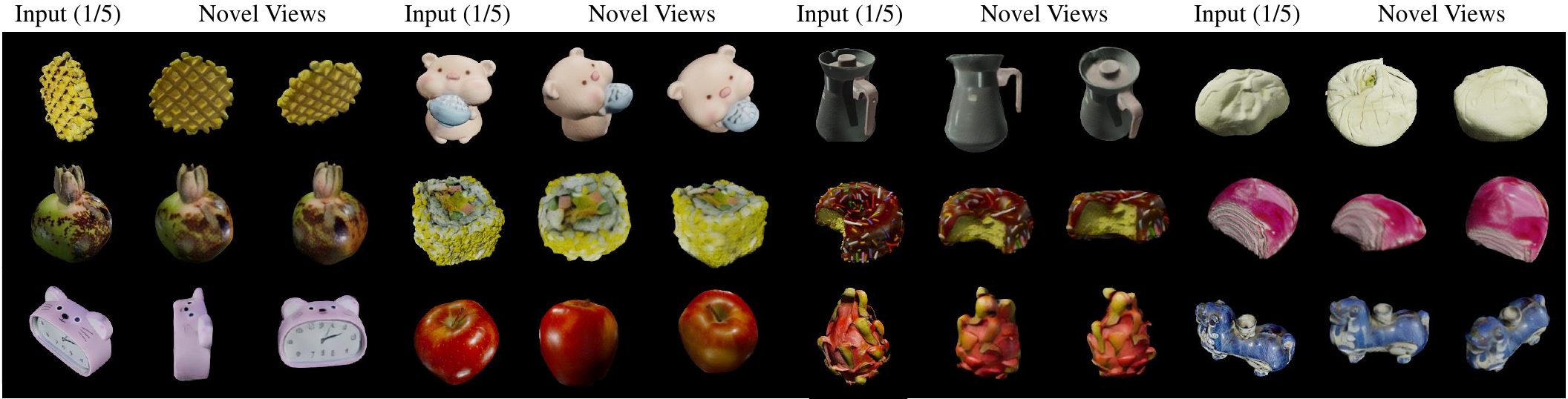}
\vspace{-0.25in}
\caption{\small{\textbf{Visualization of \modelname{} results on the object-centric Omniobject3D dataset.} For each instance, we include one out of five (denoted as ``1/5``) input views and two novel views. See more results in the supplementary.
}}
\vspace{-0.15in}
\label{fig: vis_ours}
\end{figure*}

\vspace{-2mm}
\section{Experiment}
\vspace{-2mm}

We introduce our evaluation results on diverse and large-scale datasets, including both object-centric and scene-level datasets, for comparison with prior arts.

\noindent\textbf{Implementation Details.}
We consider the number of views to be $k=5$, with image resolution $224$.
We set $\lambda_{p}=0.1$ and $\lambda_{m}=5.0$.
We set the peak learning rate as $2e-5$ (for the backbone) and $2e-4$ (for other components) with a warmup for 500 iterations using AdamW optimizer~\citep{Loshchilov2017DecoupledWD}. We train the model for 150k iterations and use a linear learning rate scheduler, where the batch size is $32$. \modelname{} has $n_e = 2$ multi-view encoder blocks and $n_m = 4$ 2D-3D mapping blocks. The resolution of the 3D neural volume and the volume-based radiance fields are $16^3$ and $64^3$, respectively. We sample $64$ points on each ray for rendering.

% \vspace{-1mm}
% \subsection{Dataset, Metrics and Baselines}
% \vspace{-1mm}

\noindent\textbf{Datasets.}
We train \modelname{} on each of the following datasets and test its capability to model the 3D object/scenes on each dataset that has different properties. We note that the poses of object-centric datasets are randomly sampled, leading to wide camera baselines of input views.

% \vspace{-0.15in}
% \paragraph{OmniObject3D dataset.} 

\vspace{-0.05in}
\hspace*{0.2in}\textbullet~~\textbf{OmniObject3D}~\citep{Wu2023OmniObject3DL3} contains daily objects from 217 categories. We use a subset with 4800 instances for training and 498 instances for testing. OmniObject3D contains objects with complicated and realistic textures.

% \vspace{-0.15in}
% \paragraph{Kubric ShapeNet dataset.} 
\vspace{-0.05in}
\hspace*{0.2in}\textbullet~~\textbf{Kubric-ShapeNet}~\citep{forge} is a synthetic dataset generated using Kubric~\citep{Greff2022KubricAS}. Its training set has 1000 instances for each of 13 ShapeNet~\citep{Chang2015ShapeNetAI} categories, resulting in $13000$ training samples. Its test set is composed of two parts: i) 1300 object instances from training categories; ii) 1000 object instances from 10 novel object categories. The two subsets are used to test the reconstruction quality and generalization ability of models.
This dataset contains objects with complicated geometry but simple textures.

% \vspace{-0.15in}
% \paragraph{Objaverse dataset.}
\vspace{-0.05in}
\hspace*{0.2in}\textbullet~~\textbf{Objaverse}~\citep{Deitke2022ObjaverseAU} is one of the largest object-centric datasets. We use a subset of $200k$ and $2k$ instances for training and testing, used to validate \modelname{} on large-scale data.

% \vspace{-0.15in}
% \paragraph{DTU dataset.}
\vspace{-0.05in}
\hspace*{0.2in}\textbullet~~\textbf{DTU dataset}~\citep{Jensen2014LargeSM} is a real scene-level dataset. DTU is on a small scale, containing only 88 scenes for training, which tests the ability of \modelname{} to fit small-scale data.

\noindent\textbf{Metrics.} Following previous works, we use standard novel view synthesis metrics, including PSNR (in dB), SSIM~\citep{Wang2004ImageQA} and LPIPS~\citep{Zhang2018TheUE}.

\noindent\textbf{Baselines.}
We compare \modelname{} with the following baselines. We note that we train each baseline model (except zero123) on each of the datasets using the same setting with \modelname{} for fair comparisons. We use official or officially verified implementations of all baselines.

\vspace{-0.05in}
\hspace*{0.2in}\textbullet~~\textbf{PixelNeRF}~\citep{pixelnerf} is a generalizable NeRF variant, using camera poses to correlate 3D points and 2D pixels. We experiment with both ground-truth poses and predicted poses (with ground-truth translations) from a state-of-the-art pose estimator RelPose~\citep{Zhang2022RelPose}.\\
\hspace*{0.2in}\textbullet~~\textbf{FORGE}~\citep{forge} is a generalizable NeRF variant with test-time optimization, which jointly predicts camera poses and the neural radiance field, and leverages their synergy to improve the performance of both. We experiment with FORGE using ground-truth and its predicted poses. \\
\hspace*{0.2in}\textbullet~~\textbf{SPARF}~\citep{Truong2022SPARFNR} is scene-specific NeRF variant that jointly optimizes the camera poses and the radiance field. It requires reasonable pose initialization and is dependent on dense visual correspondences predicted from off-the-shelf methods. \\
\hspace*{0.2in}\textbullet~~\textbf{SRT}~\citep{srt} uses only 2D representation to perform novel view synthesis. It is trained and tested on unposed image sets. \\
\hspace*{0.2in}\textbullet~~\textbf{Zero123}~\citep{zero123} is a novel view synthesis method using diffusion models.
We note that Zero123 takes a single image as input, which is different from \modelname{} and other baselines. We test Zero123 to compare \modelname{} with large-scale 2D models.

\begin{figure*}[t]
\centering
%\vspace{-0.05in}
\includegraphics[width=\linewidth]{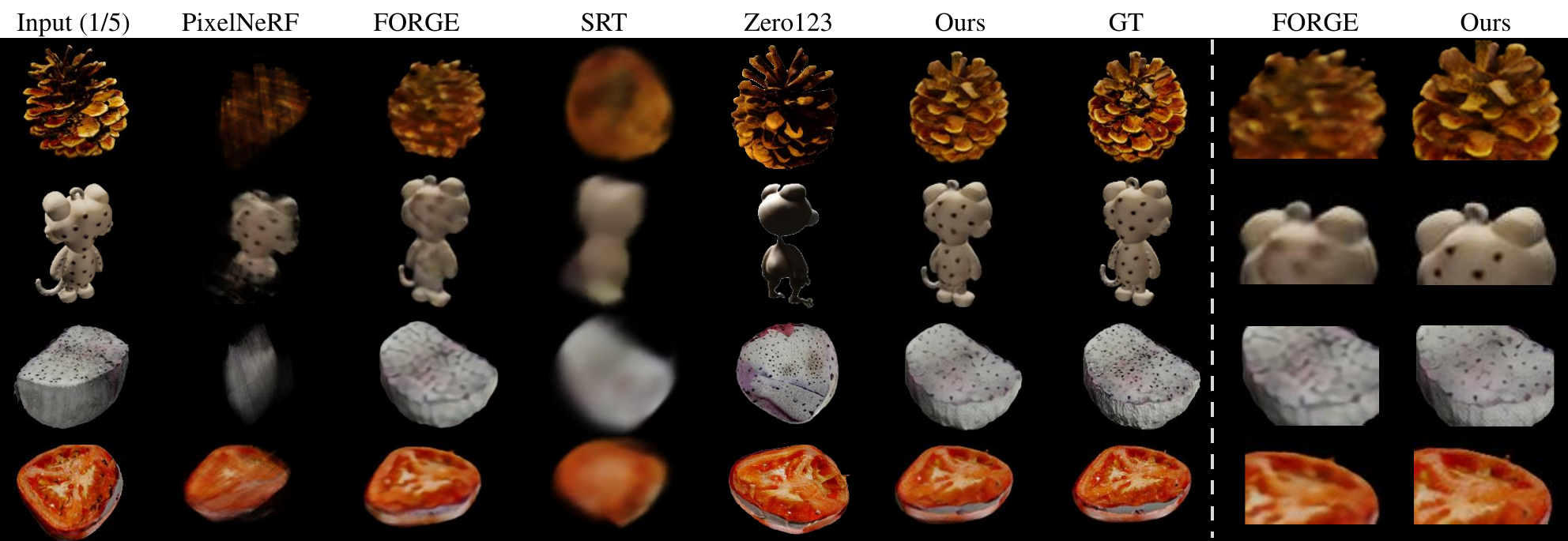}
\vspace{-0.27in}
\caption{\small{\textbf{Comparison with prior arts.} The performance of PixelNeRF degenerates dramatically with the state-of-the-art pose estimator. FORGE benefits from its joint optimization of shape and pose, but the high-frequency details are lost. SRT can only recover noisy results. Zero123 can synthesize high-quality images, while the content is not consistent with the inputs. In contrast, \modelname{} reliably recovers the details and the novel views match the ground-truth target view well. We also include zoom in results on the right for a clearer comparison.
}}
%\vspace{-0.05in}
\label{fig: vis_compare}
\end{figure*}

\begin{table*}[t]
\centering
% \tablestyle{5pt}{1.1}
\fontsize{8pt}{8pt}\selectfont
\vspace{-0.05in}
\setlength\tabcolsep{0.6pt}
\renewcommand{\arraystretch}{1.25}
\begin{tabular}{ccc|ccc|ccc|ccc|ccc}\hline
    & & &\multicolumn{3}{c|}{Omniobject3D} & \multicolumn{3}{c|}{Kubric-ShapeNet-seen} & \multicolumn{3}{c|}{Kubric-ShapeNet-novel} & 
    \multicolumn{3}{c}{Objaverse} \\

    & & &\multicolumn{3}{c|}{\textit{217 Ctg. / 498 Inst.}} & \multicolumn{3}{c|}{\textit{13 Ctg. / 1300 Inst.}} & \multicolumn{3}{c|}{\textit{10 Ctg. / 1000 Inst.}} & 
    \multicolumn{3}{c}{\textit{Open-vocab. Ctg.}} \\

    Model & Pose & Inf. Time & PSNR$_\uparrow$ & SSIM$_\uparrow$ & LPIPS$_\downarrow$ & PSNR$_\uparrow$ & SSIM$_\uparrow$ & LPIPS$_\downarrow$ & PSNR$_\uparrow$ & SSIM$_\uparrow$ & LPIPS$_\downarrow$ & PSNR$_\uparrow$ & SSIM$_\uparrow$ & LPIPS$_\downarrow$ \\ \shline
    
     \multirow{2}{*}{PixelNeRF} & GT & 2 min & 26.97 & 0.888 & 0.123 & 29.25 & 0.893 & 0.127 & 29.37 & 0.906 & 0.122 & 26.21 & 0.871 & 0.133\\

     &  Pred. & 2 min & 18.87 & 0.810 & 0.199 & 21.36 & 0.836 & 0.188 & 21.22 & 0.851 & 0.174 & 20.97 & 0.819 & 0.191\\ 
     
     \arrayrulecolor{gray}
     \cline{1-15}
     \arrayrulecolor{black}

     \multirow{2}{*}{FORGE} & GT & 0.05 sec & 28.93 & 0.913 & 0.087 & \tablegtfirst 31.32 & \tablegtfirst 0.938 & \tablegtfirst 0.053 & \tablegtfirst 31.17 & \tablegtfirst 0.946 & \tablegtfirst 0.058 & \tablegtfirst 27.76 & \tablegtfirst 0.896 & \tablegtfirst 0.100\\

     &  Pred. & 15 min & \tablesecond 26.56 & \tablesecond 0.889 & \tablesecond 0.108 & \tablesecond 26.61 & \tablesecond 0.896 & \tablesecond 0.106 & \tablesecond 25.57 & \tablesecond 0.898 & \tablesecond 0.107 & \tablesecond 23.67 & \tablesecond 0.856 & \tablesecond 0.226\\ 

     \arrayrulecolor{gray}
     \cline{1-15}
     \arrayrulecolor{black}

     SRT & \xmark & \tablesecond 0.4 sec & 20.22 & 0.786 & 0.303 & 22.62 & 0.802 & 0.267 & 22.46 & 0.793 & 0.284 & 20.41 & 0.798 & 0.312\\

     Zero123 & \xmark & 27 sec & 16.77 & 0.812 & 0.147 & 14.42 & 0.803 & 0.174 & 15.51 & 0.837 & 0.152 & 19.59 & 0.862 & 0.110\\
     
     \modelname{} & \xmark & \tablefirst \textbf{0.3 sec} & \tablefirst \textbf{29.10} & \tablefirst \textbf{0.919} & \tablefirst \textbf{0.057} & \tablefirst \textbf{29.86} & \tablefirst \textbf{0.929} & \tablefirst \textbf{0.067} & \tablefirst \textbf{28.22} & \tablefirst \textbf{0.924} & \tablefirst \textbf{0.070} & \tablefirst \textbf{26.77} & \tablefirst \textbf{0.887} & \tablefirst \textbf{0.103}\\
     \hline
    
\end{tabular}
\vspace{-0.1in}
\caption{\small{\textbf{Evaluation on four object-centric test sets.} We include the inference time of each method. \xmark\xspace means the method is pose-free. For experiments without using perfect poses, 
% We highlight the \textcolor{yellow_1}{best} and \textcolor{yellow_2}{second-best} results.
We highlight the \hlYellowOne{best} and \hlYellowTwo{second-best} results.
For experiments with perfect poses, we also highlight the \hlBlue{best GT pose result} if it is better than ours.}}
\label{tabel: main}
\vspace{-0.15in}
\end{table*}

\vspace{-1mm}
\subsection{Comparisons with State-of-the-Art}
\vspace{-1mm}
%We include both quantitative and qualitative comparisons with prior arts.

\noindent\textbf{Object-centric Results.}
The results are shown in Table~\ref{tabel: main}. 
On all four test sets, \modelname{} outperforms all prior pose-free works and pose-based works (with estimated poses). The results demonstrate the success of \modelname{} for modeling objects with different geometry and texture properties.
In detail, \modelname{} improves over the next-best baseline (FORGE) by about 3 dB PSNR and $50\%$ LPIPS relatively on all datasets. Furthermore, without the need for any test-time refinement, the running speed of \modelname{} is significantly faster than FORGE (0.3-second v.s. 15 min). Besides, \modelname{} demonstrates strong generalization capability, as the model trained on the Kubric ShapeNet dataset of only 13 categories is able to work on novel ShapeNet categories with nearly no gap.

Interestingly, when compared with prior pose-based methods using ground-truth poses, \modelname{} exhibits a comparable or even better performance. This result verifies our proposition that camera poses may not be necessary for 3D modeling, at least in the sparse-view setting. 
We present a visualization of our results in Fig.~\ref{fig: vis_ours} and comparison with prior works in Fig.~\ref{fig: vis_compare}.

\begin{figure*}[t]
\centering
\vspace{-0.1in}
\includegraphics[width=\linewidth]{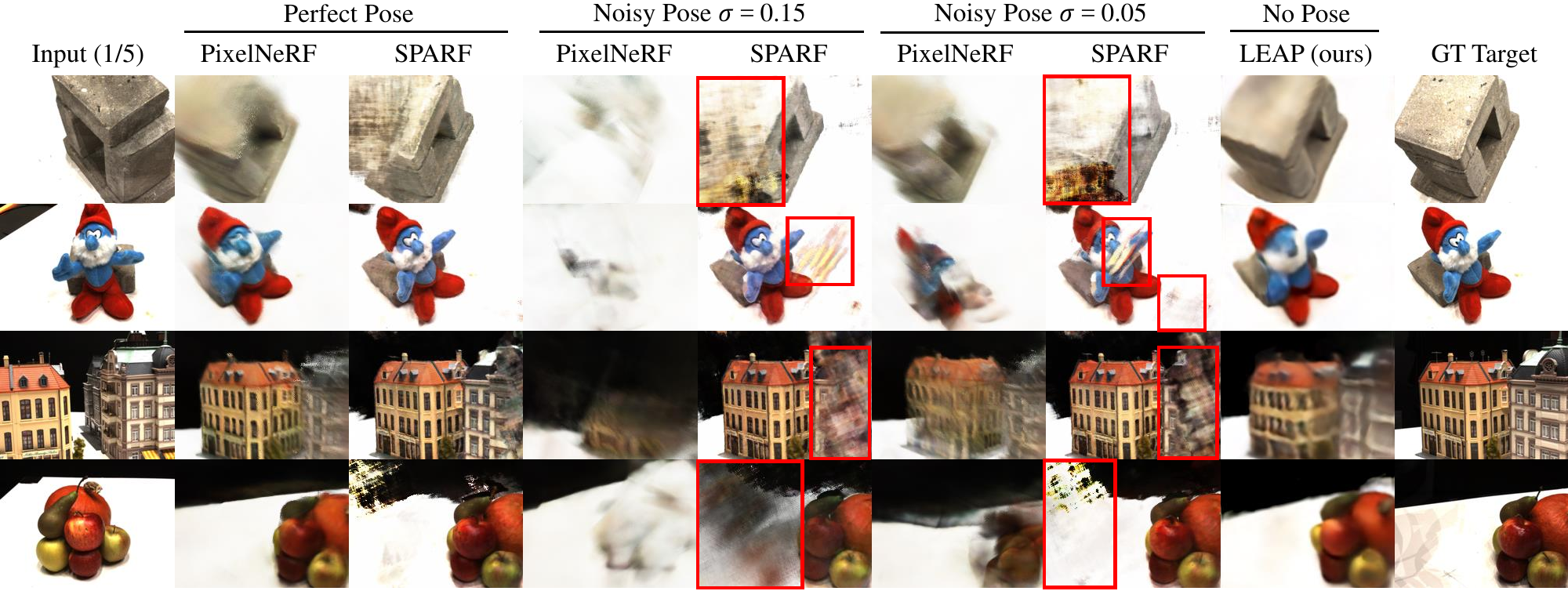}
\vspace{-0.27in}
\caption{\small{\textbf{Comparison with prior arts on DTU dataset.} PixelNeRF collapses under noisy poses. SPARF recovers high-frequency details well, but it degenerates when the correspondences are not accurate and demonstrates strong artifacts (shown in red boxes). \modelname{} reliably recovers the geometry well but lacks texture details. The result implies that \modelname{}, as a pose-free method, requires larger training datasets.
}}
%\vspace{-0.2in}
\label{fig: vis_compare_dtu}
\end{figure*}

\noindent\textbf{Scene-level Results.}
The results are shown in Table~\ref{tabel: dtu}. Since the DTU dataset is too small to train a usable pose estimator, we follow SPARF to use different levels of noisy poses.
Our method outperforms PixelNeRF with noisy poses and achieves comparable results with SPARF. We note that SPARF is a scene-specific NeRF variant that takes much longer time to optimize the radiance field and requires additional inputs, i.e. accurate dense visual correspondences between input views. We include a qualitative comparison in Fig.~\ref{fig: vis_compare_dtu}.

Besides, we also observe that a compelling phenomenon - a \modelname{} model pre-trained on the large-scale object-centric datasets largely improves its performance on the scene-level evaluation. The reason is that as a pose-free method with our geometric inductive bias, \modelname{} requires learning the knowledge on larger data compared with pose-based works. Training from scratch on the small-scale DTU dataset, which only contains 88 scenes for training, leads to unsatisfying performance. On the other hand, the effectiveness of the pre-training demonstrates the capability of \modelname{} to learn general-purpose 3D knowledge, which is generalizable and can be transferred to novel domains.

\begin{table*}[t]
% \tiny
\small
\centering
% \tablestyle{5pt}{1.1}
\fontsize{8pt}{8pt}\selectfont
\vspace{-0.05in}
\setlength\tabcolsep{2pt}
\renewcommand{\arraystretch}{1.2}
\begin{tabular}{cccccc|ccccc}\hline
    
    Method & Generalizable & Image-only & Pose Noise & Rot Err. & Trans Err. & PSNR$_\uparrow$ & SSIM$_\uparrow$ & LPIPS$_\downarrow$ & Inference Time \\ \shline

    \multirow{3}{*}{PixelNeRF} & \multirow{3}{*}{\cmark} & \multirow{3}{*}{\cmark} & GT & -- & -- & 19.60 & 0.720 & 0.295 & \multirow{3}{*}{2 min}\\

    & & & $\sigma$=0.05 & 5.03 & 0.17 & 14.42 & 0.486 & 0.463\\

    & & & $\sigma$=0.15 & 14.31 & 0.42 & 10.78 & 0.432 & 0.538\\

    \arrayrulecolor{gray}
    \cline{1-10}
     \arrayrulecolor{black}

    \multirow{3}{*}{SPARF} & \multirow{3}{*}{\xmark} & \multirow{3}{*}{\xmark} & GT & -- & -- & \tablegtfirst 19.79 & \tablegtfirst 0.749 & \tablegtfirst 0.275 & \multirow{3}{*}{1 day}\\

    & & & $\sigma$=0.05 & 1.31* & 0.04* & \tablefirst \textbf{18.57} & \tablefirst \textbf{0.682} & \tablefirst \textbf{0.336} & \\

    & & & $\sigma$=0.15 & 1.93* & 0.06* & 18.03 & 0.668 & \tablesecond 0.361 & \\

    %& & & $\sigma=0.01$ &  &  &  &  &  & \\

    \arrayrulecolor{gray}
    \cline{1-10}
    \arrayrulecolor{black}

    \modelname{} & \multirow{2}{*}{\cmark} & \multirow{2}{*}{\cmark} & \multirow{2}{*}{--} & \multirow{2}{*}{--} & \multirow{2}{*}{--} & 15.37 & 0.535 & 0.478 & \tablefirst \\

    \modelname{}-pretrain & & & & & & \tablesecond 18.07 & \tablesecond 0.671 & 0.371 & \tablefirst \multirow{-2}{*}{\textbf{0.3 sec}} \\

    \hline
\end{tabular}
\vspace{-0.1in}
\caption{\small{\textbf{Evaluation on the DTU dataset.} \modelname{} performs on par with SPARF which requires slow per-scene optimization and additional dense image correspondence inputs. Numbers with $*$ are after SPARF optimization.}}
\label{tabel: dtu}
\vspace{-0.15in}
\end{table*}

\vspace{-1mm}
\subsection{Ablation Study and Insights}
\vspace{-1mm}
We present an ablation study on each block of \modelname{} to study their impacts.

\begin{figure*}[t]
    \tiny
    \centering
    \tablestyle{5pt}{1.1}
    \begin{minipage}[t]{0.31\linewidth}
    \captionsetup{type=table}
    \tiny
    \vspace{-1.33in}
    \setlength\tabcolsep{1pt}
    \renewcommand{\arraystretch}{1.3}
    \begin{tabular}{l|ccc}\hline

     & \multicolumn{3}{c}{ShapeNet-novel} \\
    
     & PSNR$_\uparrow$ & SSIM$_\uparrow$ & LPIPS$_\downarrow$ \\ \shline

    \textbf{\modelname{}} (full) & \textbf{28.22} & \textbf{0.924} & \textbf{0.072}\\ \hline

    a) use world frame &  24.62 & 0.865 & 0.116\\

    \arrayrulecolor{gray}
    \cline{1-4}
    \arrayrulecolor{black}

    b) no multi-view enc. & 21.68 & 0.710 & 0.431\\

    c) only GCR layer & 22.98 & 0.770 & 0.359\\

    d) only NVU layer & 27.62 & 0.907 & 0.085 \\

    \arrayrulecolor{gray}
    \cline{1-4}
    \arrayrulecolor{black}

    e) two mapping layers  & 27.99 & 0.910 & 0.080\\

    \hline
    \end{tabular}
    \vspace{-0.1in}
    \caption{\small{\textbf{Ablation study on the Kubric dataset.} We ablate on a) using category-level world canonical space rather than the camera space. b)-d) the multi-view encoder designs for enabling the model aware of canonical view choice. And e) using less ($\#$2/4) mapping layers. See visualization in supplementary.
    }}
    \label{tabel: ablation}
    \end{minipage}
    \hfill
    \begin{minipage}[t]{0.65\linewidth}
    \centering
    \includegraphics[width=0.97\linewidth]{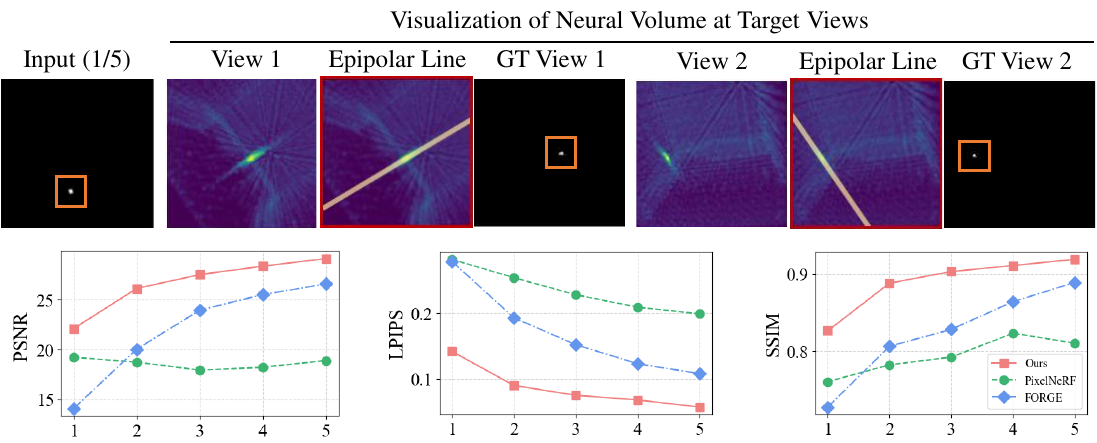}
    \vspace{-0.1in}
    \caption{\small{\textbf{Interpret \modelname{}.} Top: We input images of a small dot (in orange boxes), and the visualization of the reconstructed neural volume shows consistency with the epipolar lines of the small dot on target views. This implies \modelname{} mapps a 2D point as its 3D reprojection ray segment even though there are no reprojection operations. It leverages the multi-view information to resolve the depth ambiguity of the ray. Bottom: The performance with different numbers of inputs on Omniobject3D. Note that we only train the model with 5 images and it is directly tested.} 
    }
    \label{fig: epipolar}
    \end{minipage}
\vspace{-0.15in}
\end{figure*}

\noindent\textbf{Coordinate Frame.} We study the importance of the local camera coordinate frame by using the world coordinate frame instead. We use the category-level coordinate as the world coordinate, where objects have aligned rotation and are zero-centered. As shown in Table~\ref{tabel: ablation} (a), the model demonstrates better performance on the seen object categories (with 31.23 v.s. 29.10 PSNR) but generalizes worse on novel categories. We conjecture the reason is that the aligned rotation/translation world coordinate frame makes it easier to perform 2D-3D information mapping for training categories. However, it also limits the performance of novel categories, as their category-level coordinate frames are not learned by \modelname{}. This result matches our intuition of using the local camera coordinate frame to define the neural volume, which enables \modelname{} to generalize to any objects/scenes.

\begin{figure*}[t]
\centering
%\vspace{-0.1in}
\includegraphics[width=\linewidth]{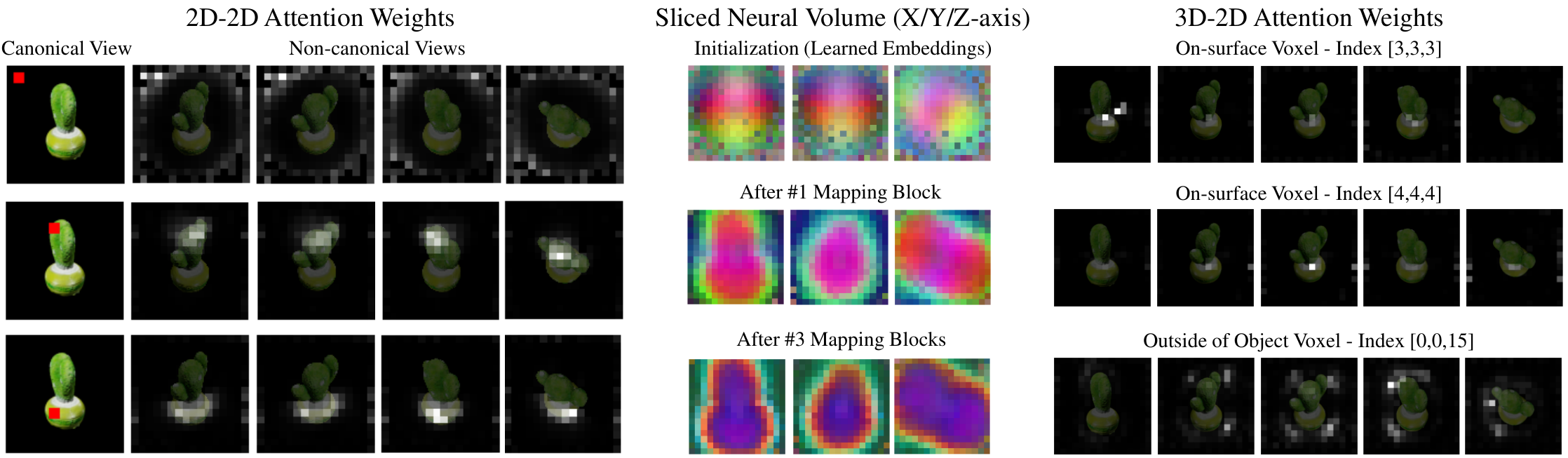}
\vspace{-0.23in}
\caption{\small{\textbf{Visualization of \modelname{} working mechanism.} (Left) We show the 2D-2D attention weights of the multi-view encoder. For each query pixel (in red) in the canonical view, it assigns larger weights to the corresponding regions in the Non-canonical views. The attention of query points in the background diffuses. (Middle) We visualize the learned neural volume by using PCA and slicing along the three axes. As our 3D modeling happens in the local coordinate frame, which is not axis-aligned, the learned embeddings show isotropic properties. The neural volume is refined in a coarse-to-fine manner, where the object boundary becomes more compact after more mapping layers. (Right) We show the attention map between 3D-2D. As shown in the top two rows, the neighbor on-surface voxels have similar attention patterns on a specific 2D object region. The attention of an out-of-object voxel diffuses on the background 2D region. See more details in the supplementary.
}}
\vspace{-0.1in}
\label{fig: vis_attn_main}
\end{figure*}

\noindent\textbf{Multi-view Encoder.} We explore the impact of using the multi-view encoder to make \modelname{} aware of the choice of the canonical view. We test the following alternatives: i) \modelname{} without the multi-view encoder; ii) \modelname{} that only has the global consensus reasoning (GCR) layers; iii) \modelname{} that only has the non-canonical view update (NVU) layers. As shown in Table~\ref{tabel: ablation} (b)-(d), without the multi-view encoder, we observe a significant performance drop. The reason is that the inconsistent features across views hamper the 2D-3D information mapping. Similarly, only with the GCR layers, \modelname{} struggles determining which view is the canonical view. When only using the NVU layers, it achieves a slightly worse performance than the full model. The experiments show the effectiveness of using the multi-view encoder to make the model aware of the choice of canonical view.

\noindent\textbf{The 2D-3D Information Mapping Layer.} As shown in Table~\ref{tabel: ablation} (e), using two mapping layers (the default has four layers) slightly degenerates the performance, which shows its efficacy.

\noindent\textbf{Interpreting \modelname{}.} We perform visualization to understand what knowledge \modelname{} learns to handle the absence of camera poses. As shown in Fig.~\ref{fig: vis_attn_main}, \modelname{} adaptively assigns weights to reasonable 2D regions to perform 2D-2D reasoning and 2D-3D information mapping.
The neural volume is updated in a coarse-to-fine manner during the process. Moreover, we test how the learned knowledge is related to explicit pose-based operations. 
As shown in Fig.~\ref{fig: epipolar}, we input images of a small dot.
We find that \modelname{} lifts the 2D pixel of the dot into the 3D space as a line segment. The location of the line segment projected in another view corresponds to its epipolar line. The phenomenon reveals that \modelname{} lifts a 2D point as its reprojection ray, and it leverages multiview information to resolve the ambiguity of the ray to determine the depth of the 2D point. Besides, we show \modelname{} performance with different numbers of input images. The results show that \modelname{} reliably reconstructs the object with two to five images, and its performance drops slightly with fewer inputs. However, we observe a big drop when we decrease the number of inputs from two images to one image. These results validate the effectiveness of \modelname{} in using multi-view information to perform the 3D modeling. %\textbf{We include more analyses in the supplementary to show the visualization of 2D-2D and 3D-2D attention weights, as well as the learned neural volume.}

\vspace{-2mm}
\section{Conclusion}
\vspace{-2mm}

We propose \modelname{}, a pose-free approach for 3D modeling from a set of unposed sparse-view images. By appropriately setting the 3D coordinate and aggregating 2D image features, \modelname{} demonstrates satisfying novel view synthesis quality. In our experiments, spanning from both object-centric to scene-level, from synthetic images to real images, and from small-scale to large-scale data, \modelname{} consistently demonstrates better performance compared with prior pose-based works that use estimated poses or noisy poses. \modelname{} also achieve comparable results with the versions of prior works that use ground-truth poses. Besides, \modelname{} showcases a strong generalization capability, fast inference speed, and interpretable learned knowledge.

\noindent\textbf{Limitations.} \modelname{} adopts the neural volume representation where the 3D voxel grids span uniformly in the 3D space, and the physical size of the volume is bounded. Designing better 3D representation, e.g. incorporating techniques from prior works to enable it to work on unbounded scenes, will further benefit the application of \modelname{}.

% \subsubsection*{Acknowledgments}
% Use unnumbered third level headings for the acknowledgments. All
% acknowledgments, including those to funding agencies, go at the end of the paper.

\bibliography{egbib}

\begin{thebibliography}{43}
\providecommand{\natexlab}[1]{#1}
\providecommand{\url}[1]{\texttt{#1}}
\expandafter\ifx\csname urlstyle\endcsname\relax
  \providecommand{\doi}[1]{doi: #1}\else
  \providecommand{\doi}{doi: \begingroup \urlstyle{rm}\Url}\fi

\bibitem[Caron et~al.(2021)Caron, Touvron, Misra, J'egou, Mairal, Bojanowski, and Joulin]{Caron2021EmergingPI}
Mathilde Caron, Hugo Touvron, Ishan Misra, Herv'e J'egou, Julien Mairal, Piotr Bojanowski, and Armand Joulin.
\newblock Emerging properties in self-supervised vision transformers.
\newblock \emph{2021 IEEE/CVF International Conference on Computer Vision (ICCV)}, pp.\  9630--9640, 2021.

\bibitem[Chang et~al.(2015)Chang, Funkhouser, Guibas, Hanrahan, Huang, Li, Savarese, Savva, Song, Su, Xiao, Yi, and Yu]{Chang2015ShapeNetAI}
Angel~X. Chang, Thomas~A. Funkhouser, Leonidas~J. Guibas, Pat Hanrahan, Qixing Huang, Zimo Li, Silvio Savarese, Manolis Savva, Shuran Song, Hao Su, Jianxiong Xiao, L.~Yi, and Fisher Yu.
\newblock Shapenet: An information-rich 3d model repository.
\newblock \emph{ArXiv}, abs/1512.03012, 2015.

\bibitem[Chen et~al.(2021{\natexlab{a}})Chen, Xu, Zhao, Zhang, Xiang, Yu, and Su]{Chen2021MVSNeRF}
Anpei Chen, Zexiang Xu, Fuqiang Zhao, Xiaoshuai Zhang, Fanbo Xiang, Jingyi Yu, and Hao Su.
\newblock Mvsnerf: Fast generalizable radiance field reconstruction from multi-view stereo.
\newblock \emph{ICCV}, pp.\  14104--14113, 2021{\natexlab{a}}.

\bibitem[Chen et~al.(2021{\natexlab{b}})Chen, Snavely, and Makadia]{chen2021widebaseline1}
Kefan Chen, Noah Snavely, and Ameesh Makadia.
\newblock Wide-baseline relative camera pose estimation with directional learning.
\newblock In \emph{Proceedings of the IEEE/CVF Conference on Computer Vision and Pattern Recognition}, pp.\  3258--3268, 2021{\natexlab{b}}.

\bibitem[Deitke et~al.(2022)Deitke, Schwenk, Salvador, Weihs, Michel, VanderBilt, Schmidt, Ehsani, Kembhavi, and Farhadi]{Deitke2022ObjaverseAU}
Matt Deitke, Dustin Schwenk, Jordi Salvador, Luca Weihs, Oscar Michel, Eli VanderBilt, Ludwig Schmidt, Kiana Ehsani, Aniruddha Kembhavi, and Ali Farhadi.
\newblock Objaverse: A universe of annotated 3d objects.
\newblock \emph{ArXiv}, abs/2212.08051, 2022.

\bibitem[Deng et~al.(2021)Deng, Litany, Duan, Poulenard, Tagliasacchi, and Guibas]{Deng2021VectorNerons}
Congyue Deng, Or~Litany, Yueqi Duan, Adrien Poulenard, Andrea Tagliasacchi, and Leonidas~J. Guibas.
\newblock Vector neurons: A general framework for so(3)-equivariant networks.
\newblock \emph{ICCV}, pp.\  12180--12189, 2021.

\bibitem[Dosovitskiy et~al.(2020)Dosovitskiy, Beyer, Kolesnikov, Weissenborn, Zhai, Unterthiner, Dehghani, Minderer, Heigold, Gelly, Uszkoreit, and Houlsby]{vit}
Alexey Dosovitskiy, Lucas Beyer, Alexander Kolesnikov, Dirk Weissenborn, Xiaohua Zhai, Thomas Unterthiner, Mostafa Dehghani, Matthias Minderer, Georg Heigold, Sylvain Gelly, Jakob Uszkoreit, and Neil Houlsby.
\newblock An image is worth 16x16 words: Transformers for image recognition at scale.
\newblock \emph{ArXiv}, abs/2010.11929, 2020.

\bibitem[Geiger et~al.(2011)Geiger, Ziegler, and Stiller]{Geiger2011StereoScanD3}
Andreas Geiger, Julius Ziegler, and Christoph Stiller.
\newblock Stereoscan: Dense 3d reconstruction in real-time.
\newblock \emph{2011 IEEE Intelligent Vehicles Symposium (IV)}, pp.\  963--968, 2011.

\bibitem[Goesele et~al.(2006)Goesele, Curless, and Seitz]{goesele2006multi}
Michael Goesele, Brian Curless, and Steven~M Seitz.
\newblock Multi-view stereo revisited.
\newblock In \emph{CVPR}, volume~2, pp.\  2402--2409. IEEE, 2006.

\bibitem[Greff et~al.(2022)Greff, Belletti, Beyer, Doersch, Du, Duckworth, Fleet, Gnanapragasam, Golemo, Herrmann, Kipf, Kundu, Lagun, Laradji, Liu, Meyer, Miao, Nowrouzezahrai, Oztireli, Pot, Radwan, Rebain, Sabour, Sajjadi, Sela, Sitzmann, Stone, Sun, Vora, Wang, Wu, Yi, Zhong, and Tagliasacchi]{Greff2022KubricAS}
Klaus Greff, Francois Belletti, Lucas Beyer, Carl Doersch, Yilun Du, Daniel Duckworth, David~J. Fleet, Dan Gnanapragasam, Florian Golemo, Charles Herrmann, Thomas Kipf, Abhijit Kundu, Dmitry Lagun, Issam~Hadj Laradji, Hsueh-Ti Liu, Henning Meyer, Yishu Miao, Derek Nowrouzezahrai, Cengiz Oztireli, Etienne Pot, Noha Radwan, Daniel Rebain, Sara Sabour, Mehdi S.~M. Sajjadi, Matan Sela, Vincent Sitzmann, Austin Stone, Deqing Sun, Suhani Vora, Ziyu Wang, Tianhao Wu, Kwang~Moo Yi, Fangcheng Zhong, and Andrea Tagliasacchi.
\newblock Kubric: A scalable dataset generator.
\newblock \emph{2022 IEEE/CVF Conference on Computer Vision and Pattern Recognition (CVPR)}, pp.\  3739--3751, 2022.

\bibitem[Hutchcroft et~al.(2022)Hutchcroft, Li, Boyadzhiev, Wan, Wang, and Kang]{hutchcroft2022covispose}
Will Hutchcroft, Yuguang Li, Ivaylo Boyadzhiev, Zhiqiang Wan, Haiyan Wang, and Sing~Bing Kang.
\newblock Covispose: Co-visibility pose transformer for wide-baseline relative pose estimation in 360° indoor panoramas.
\newblock In \emph{European Conference on Computer Vision}, pp.\  615--633. Springer, 2022.

\bibitem[Jensen et~al.(2014)Jensen, Dahl, Vogiatzis, Tola, and Aan{\ae}s]{Jensen2014LargeSM}
Rasmus~Ramsb{\o}l Jensen, A.~Dahl, George Vogiatzis, Engil Tola, and Henrik Aan{\ae}s.
\newblock Large scale multi-view stereopsis evaluation.
\newblock \emph{CVPR}, pp.\  406--413, 2014.

\bibitem[Jiang et~al.(2022)Jiang, Jiang, Grauman, and Zhu]{forge}
Hanwen Jiang, Zhenyu Jiang, Kristen Grauman, and Yuke Zhu.
\newblock Few-view object reconstruction with unknown categories and camera poses.
\newblock \emph{ArXiv}, abs/2212.04492, 2022.

\bibitem[Johnson et~al.(2016)Johnson, Alahi, and Fei-Fei]{perceptualLoss}
Justin Johnson, Alexandre Alahi, and Li~Fei-Fei.
\newblock Perceptual losses for real-time style transfer and super-resolution.
\newblock In \emph{ECCV}, 2016.

\bibitem[Lin et~al.(2023)Lin, Zhang, Ramanan, and Tulsiani]{relpose2}
Amy Lin, Jason~Y. Zhang, Deva Ramanan, and Shubham Tulsiani.
\newblock Relpose++: Recovering 6d poses from sparse-view observations.
\newblock \emph{ArXiv}, abs/2305.04926, 2023.

\bibitem[Lin et~al.(2021)Lin, Ma, Torralba, and Lucey]{Lin2021BARFBN}
Chen-Hsuan Lin, Wei-Chiu Ma, Antonio Torralba, and Simon Lucey.
\newblock Barf: Bundle-adjusting neural radiance fields.
\newblock \emph{2021 IEEE/CVF International Conference on Computer Vision (ICCV)}, pp.\  5721--5731, 2021.

\bibitem[Liu et~al.(2023)Liu, Wu, Hoorick, Tokmakov, Zakharov, and Vondrick]{zero123}
Ruoshi Liu, Rundi Wu, Basile~Van Hoorick, Pavel Tokmakov, Sergey Zakharov, and Carl Vondrick.
\newblock Zero-1-to-3: Zero-shot one image to 3d object.
\newblock \emph{ICCV}, 2023.

\bibitem[Loshchilov \& Hutter(2017)Loshchilov and Hutter]{Loshchilov2017DecoupledWD}
Ilya Loshchilov and Frank Hutter.
\newblock Decoupled weight decay regularization.
\newblock In \emph{International Conference on Learning Representations}, 2017.

\bibitem[Mildenhall et~al.(2020)Mildenhall, Srinivasan, Tancik, Barron, Ramamoorthi, and Ng]{Mildenhall2020NeRFRS}
Ben Mildenhall, Pratul~P. Srinivasan, Matthew Tancik, Jonathan~T. Barron, Ravi Ramamoorthi, and Ren Ng.
\newblock Nerf: Representing scenes as neural radiance fields for view synthesis.
\newblock \emph{ArXiv}, abs/2003.08934, 2020.

\bibitem[Niemeyer et~al.(2022)Niemeyer, Barron, Mildenhall, Sajjadi, Geiger, and Radwan]{Niemeyer2021RegNeRF}
Michael Niemeyer, Jonathan~T. Barron, Ben Mildenhall, Mehdi S.~M. Sajjadi, Andreas Geiger, and Noha Radwan.
\newblock Regnerf: Regularizing neural radiance fields for view synthesis from sparse inputs.
\newblock \emph{CVPR}, pp.\  5470--5480, 2022.

\bibitem[Oquab et~al.(2023)Oquab, Darcet, Moutakanni, Vo, Szafraniec, Khalidov, Fernandez, Haziza, Massa, El-Nouby, Assran, Ballas, Galuba, Howes, Huang, Li, Misra, Rabbat, Sharma, Synnaeve, Xu, J{\'e}gou, Mairal, Labatut, Joulin, and Bojanowski]{Oquab2023DINOv2LR}
Maxime Oquab, Timoth'ee Darcet, Th'eo Moutakanni, Huy~Q. Vo, Marc Szafraniec, Vasil Khalidov, Pierre Fernandez, Daniel Haziza, Francisco Massa, Alaaeldin El-Nouby, Mahmoud Assran, Nicolas Ballas, Wojciech Galuba, Russ Howes, Po-Yao~(Bernie) Huang, Shang-Wen Li, Ishan Misra, Michael~G. Rabbat, Vasu Sharma, Gabriel Synnaeve, Huijiao Xu, Herv{\'e} J{\'e}gou, Julien Mairal, Patrick Labatut, Armand Joulin, and Piotr Bojanowski.
\newblock Dinov2: Learning robust visual features without supervision.
\newblock \emph{ArXiv}, abs/2304.07193, 2023.

\bibitem[Qi et~al.(2017)Qi, Su, Mo, and Guibas]{Qi2016PointNet}
C.~Qi, Hao Su, Kaichun Mo, and Leonidas~J. Guibas.
\newblock Pointnet: Deep learning on point sets for 3d classification and segmentation.
\newblock \emph{CVPR}, pp.\  77--85, 2017.

\bibitem[Rockwell et~al.(2022)Rockwell, Johnson, and Fouhey]{rockwell202288point}
Chris Rockwell, Justin Johnson, and David~F Fouhey.
\newblock The 8-point algorithm as an inductive bias for relative pose prediction by vits.
\newblock In \emph{2022 International Conference on 3D Vision (3DV)}, pp.\  1--11. IEEE, 2022.

\bibitem[Sajjadi et~al.(2022)Sajjadi, Meyer, Pot, Bergmann, Greff, Radwan, Vora, Lucic, Duckworth, Dosovitskiy, Uszkoreit, Funkhouser, and Tagliasacchi]{srt}
Mehdi S.~M. Sajjadi, Henning Meyer, Etienne Pot, Urs~M. Bergmann, Klaus Greff, Noha Radwan, Suhani Vora, Mario Lucic, Daniel Duckworth, Alexey Dosovitskiy, Jakob Uszkoreit, Thomas~A. Funkhouser, and Andrea Tagliasacchi.
\newblock Scene representation transformer: Geometry-free novel view synthesis through set-latent scene representations.
\newblock \emph{CVPR}, 2022.

\bibitem[Sch\"{o}nberger \& Frahm(2016)Sch\"{o}nberger and Frahm]{schoenberger2016sfm}
Johannes~Lutz Sch\"{o}nberger and Jan-Michael Frahm.
\newblock {Structure-from-Motion Revisited}.
\newblock In \emph{Conference on Computer Vision and Pattern Recognition (CVPR)}, 2016.

\bibitem[Sinha et~al.(2022)Sinha, Zhang, Tagliasacchi, Gilitschenski, and Lindell]{sparsepose}
Samarth Sinha, Jason~Y. Zhang, Andrea Tagliasacchi, Igor Gilitschenski, and David~B. Lindell.
\newblock Sparsepose: Sparse-view camera pose regression and refinement.
\newblock \emph{ArXiv}, abs/2211.16991, 2022.

\bibitem[Sun et~al.(2021)Sun, Shen, Wang, Bao, and Zhou]{sun2021loftr}
Jiaming Sun, Zehong Shen, Yuang Wang, Hujun Bao, and Xiaowei Zhou.
\newblock Loftr: Detector-free local feature matching with transformers.
\newblock In \emph{Proceedings of the IEEE/CVF conference on computer vision and pattern recognition}, pp.\  8922--8931, 2021.

\bibitem[Truong et~al.(2022)Truong, Rakotosaona, Manhardt, and Tombari]{Truong2022SPARFNR}
Prune Truong, Marie-Julie Rakotosaona, Fabian Manhardt, and Federico Tombari.
\newblock Sparf: Neural radiance fields from sparse and noisy poses.
\newblock \emph{ArXiv}, abs/2211.11738, 2022.

\bibitem[Vaswani et~al.(2017)Vaswani, Shazeer, Parmar, Uszkoreit, Jones, Gomez, Kaiser, and Polosukhin]{transformer}
Ashish Vaswani, Noam~M. Shazeer, Niki Parmar, Jakob Uszkoreit, Llion Jones, Aidan~N. Gomez, Lukasz Kaiser, and Illia Polosukhin.
\newblock Attention is all you need.
\newblock In \emph{NIPS}, 2017.

\bibitem[Wang et~al.(2021{\natexlab{a}})Wang, Liu, Liu, Theobalt, Komura, and Wang]{Wang2021NeuSLN}
Peng Wang, Lingjie Liu, Yuan Liu, Christian Theobalt, Taku Komura, and Wenping Wang.
\newblock Neus: Learning neural implicit surfaces by volume rendering for multi-view reconstruction.
\newblock In \emph{NeurIPS}, 2021{\natexlab{a}}.

\bibitem[Wang et~al.(2021{\natexlab{b}})Wang, Wang, Genova, Srinivasan, Zhou, Barron, Martin-Brualla, Snavely, and Funkhouser]{Wang2021IBRNet}
Qianqian Wang, Zhicheng Wang, Kyle Genova, Pratul~P. Srinivasan, Howard Zhou, Jonathan~T. Barron, Ricardo Martin-Brualla, Noah Snavely, and Thomas~A. Funkhouser.
\newblock Ibrnet: Learning multi-view image-based rendering.
\newblock \emph{CVPR}, pp.\  4688--4697, 2021{\natexlab{b}}.

\bibitem[Wang et~al.(2004)Wang, Bovik, Sheikh, and Simoncelli]{Wang2004ImageQA}
Zhou Wang, Alan~Conrad Bovik, Hamid~R. Sheikh, and Eero~P. Simoncelli.
\newblock Image quality assessment: from error visibility to structural similarity.
\newblock \emph{IEEE Transactions on Image Processing}, 13:\penalty0 600--612, 2004.

\bibitem[Wang et~al.(2021{\natexlab{c}})Wang, Wu, Xie, Chen, and Prisacariu]{Wang2021NeRF--}
Zirui Wang, Shangzhe Wu, Weidi Xie, Min Chen, and Victor~Adrian Prisacariu.
\newblock Nerf-: Neural radiance fields without known camera parameters.
\newblock \emph{ArXiv}, abs/2102.07064, 2021{\natexlab{c}}.

\bibitem[Wu et~al.(2023)Wu, Zhang, Fu, Wang, Ren, Pan, Wu, Yang, Wang, Qian, Lin, and Liu]{Wu2023OmniObject3DL3}
Tong Wu, Jiarui Zhang, Xiao Fu, Yuxin Wang, Jiawei Ren, Liang Pan, Wayne Wu, Lei Yang, Jiaqi Wang, Chen Qian, Dahua Lin, and Ziwei Liu.
\newblock Omniobject3d: Large-vocabulary 3d object dataset for realistic perception, reconstruction and generation.
\newblock \emph{CVPR}, 2023.

\bibitem[Xia et~al.(2022)Xia, Tang, Timofte, and Gool]{Xia2022SiNeRFSN}
Yitong Xia, Hao Tang, Radu Timofte, and Luc~Van Gool.
\newblock Sinerf: Sinusoidal neural radiance fields for joint pose estimation and scene reconstruction.
\newblock In \emph{British Machine Vision Conference}, 2022.

\bibitem[Yang et~al.(2023)Yang, Pavone, and Wang]{Yang2023FreeNeRF}
Jiawei Yang, Marco Pavone, and Yue Wang.
\newblock Freenerf: Improving few-shot neural rendering with free frequency regularization.
\newblock \emph{CVPR}, 2023.

\bibitem[Yu et~al.(2021)Yu, Ye, Tancik, and Kanazawa]{pixelnerf}
Alex Yu, Vickie Ye, Matthew Tancik, and Angjoo Kanazawa.
\newblock pixelnerf: Neural radiance fields from one or few images.
\newblock \emph{CVPR}, pp.\  4576--4585, 2021.

\bibitem[Yu et~al.(2022)Yu, Fridovich-Keil, Tancik, Chen, Recht, and Kanazawa]{Yu2021PlenoxelsRF}
Alex Yu, Sara Fridovich-Keil, Matthew Tancik, Qinhong Chen, Benjamin Recht, and Angjoo Kanazawa.
\newblock Plenoxels: Radiance fields without neural networks.
\newblock \emph{cvpr}, pp.\  5491--5500, 2022.

\bibitem[Zhang et~al.(2021)Zhang, Yang, Tulsiani, and Ramanan]{zhang2021ners}
Jason~Y. Zhang, Gengshan Yang, Shubham Tulsiani, and Deva Ramanan.
\newblock {NeRS}: Neural reflectance surfaces for sparse-view 3d reconstruction in the wild.
\newblock In \emph{Conference on Neural Information Processing Systems}, 2021.

\bibitem[Zhang et~al.(2022)Zhang, Ramanan, and Tulsiani]{Zhang2022RelPose}
Jason~Y. Zhang, Deva Ramanan, and Shubham Tulsiani.
\newblock Relpose: Predicting probabilistic relative rotation for single objects in the wild.
\newblock 2022.

\bibitem[Zhang et~al.(2023)Zhang, Herrmann, Hur, Cabrera, Jampani, Sun, and Yang]{Zhang2023Ataleoftwofeat}
Junyi Zhang, Charles Herrmann, Junhwa Hur, Luisa~Polania Cabrera, Varun Jampani, Deqing Sun, and Ming Yang.
\newblock A tale of two features: Stable diffusion complements dino for zero-shot semantic correspondence.
\newblock \emph{ArXiv}, abs/2305.15347, 2023.

\bibitem[Zhang et~al.(2018)Zhang, Isola, Efros, Shechtman, and Wang]{Zhang2018TheUE}
Richard Zhang, Phillip Isola, Alexei~A. Efros, Eli Shechtman, and Oliver Wang.
\newblock The unreasonable effectiveness of deep features as a perceptual metric.
\newblock \emph{2018 IEEE/CVF Conference on Computer Vision and Pattern Recognition}, pp.\  586--595, 2018.

\bibitem[Zisserman(2001)]{Zisserman2001MultipleVG}
Andrew Zisserman.
\newblock Multiple view geometry in computer vision.
\newblock \emph{K{\"u}nstliche Intell.}, 15:\penalty0 41, 2001.

\end{thebibliography}
\bibliographystyle{iclr2024_conference}

\appendix
\clearpage
\appendixpage
In the supplementary, we first include a detailed analysis to understand the novel pose-free framework of \modelname{} by visualizing the intermediate results. Then, we include more novel view synthesis visualization results. Finally, we include the details of our model and the baselines, as well as the evaluation details on each dataset.

\section{More Analysis}
In this section, we dive deep into the pose-free framework and visualize the intermediate layer results and attention weights to understand the working mechanism of \modelname{}. Following the sequence of modules of \modelname{}, we perform analyses on the Multi-view Encoder and the 2D-to-3D information mapping module.

\begin{figure*}[h]
\centering
\vspace{-0.1in}
\includegraphics[width=0.7\linewidth]{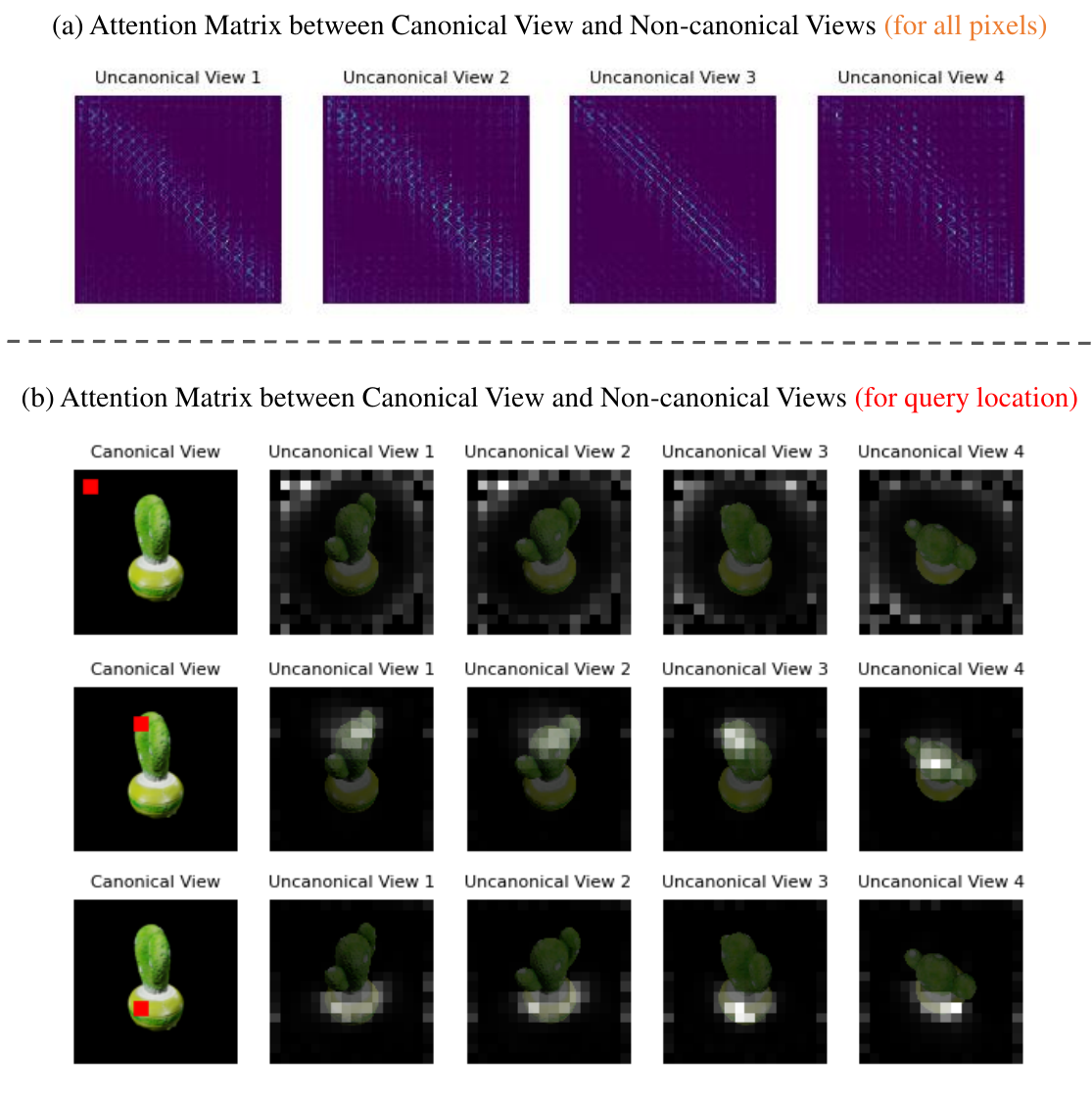}
\vspace{-0.15in}
\caption{\small{\textbf{Visualization of cross-attention weight in the Non-canonical View Update layer in the Multi-view Encoder.} (a) We show the attention matrix on the entire image, where the shape of the attention matrix is $\mathbb{R}^{hw\times hw}$ and $h$, $w$ are the resolution of the image features. The visualization demonstrates clear patterns of the correlation between canonical view and non-canonical view image features. (b) We further show detailed attention weights for each query pixel in the canonical view. The visualization demonstrates that the attention captures the cross-view correlations well. In detail, for a query pixel in the background, the corresponding regions are also in the background of the non-canonical views. When the query pixel is located on the object, the attention focuses on small corresponding regions in the non-canonical views. The results demonstrate that the image features of the multi-view images are coherent.
}}
%\vspace{-0.2in}
\label{fig: analysis_2d_attn}
\end{figure*}

\paragraph{Multi-view Encoder.} We first understand the Non-canonical View Update layers. We visualize the cross-attention weights between the canonical views and the non-canonical views. The results are shown in Fig.~\ref{fig: analysis_2d_attn}.

\begin{figure*}[h]
\centering
% \vspace{-0.3in}
\includegraphics[width=\linewidth]{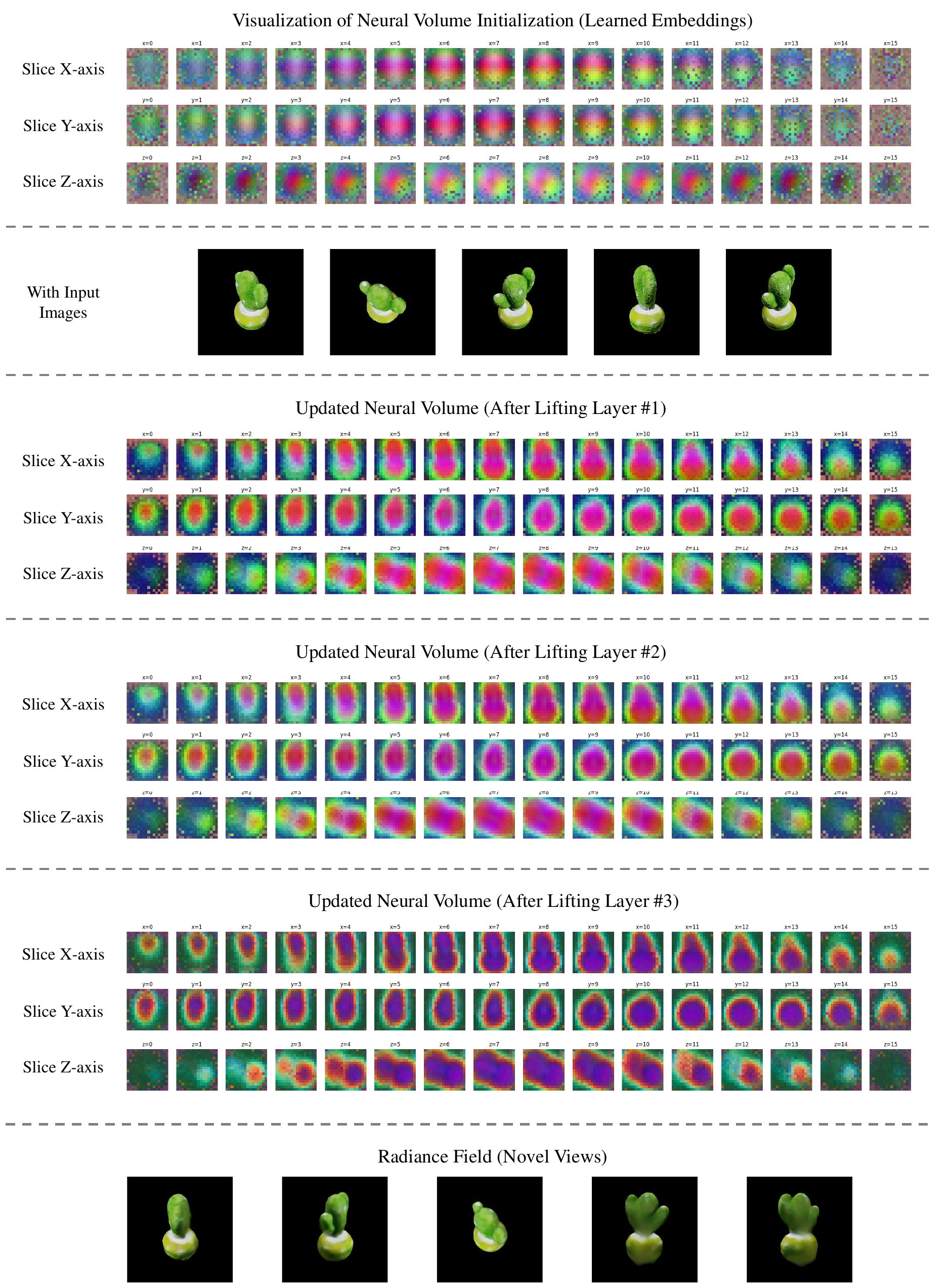}
\vspace{-0.1in}
\caption{\small{\textbf{Visualization of learned neural volume embeddings and the updated neural volume.} We visualize the neural volume by slicing it from the three axes.
(Top) As our 3D modeling happens in the local camera coordinate which is not axis-aligned, the learned embeddings show isotropic properties, i.e. like a sphere which is able to model objects with different orientations. (Bottom) After lifting information from the images, the neural volume is turned into the shape of the input object. With more 2D-to-3D information mapping layers, the boundary of the object becomes more sharp.
}}
\vspace{-0.2in}
\label{fig: analysis_embeddings}
\end{figure*}

\paragraph{The Learned Neural Volume Embeddings.} We then visualize the learned neural volume embeddings and the updated neural volume through the 2D-to-3D information mapping layers. The results are shown in Fig.~\ref{fig: analysis_embeddings}.

\begin{figure*}[t]
\centering
\includegraphics[width=\linewidth]{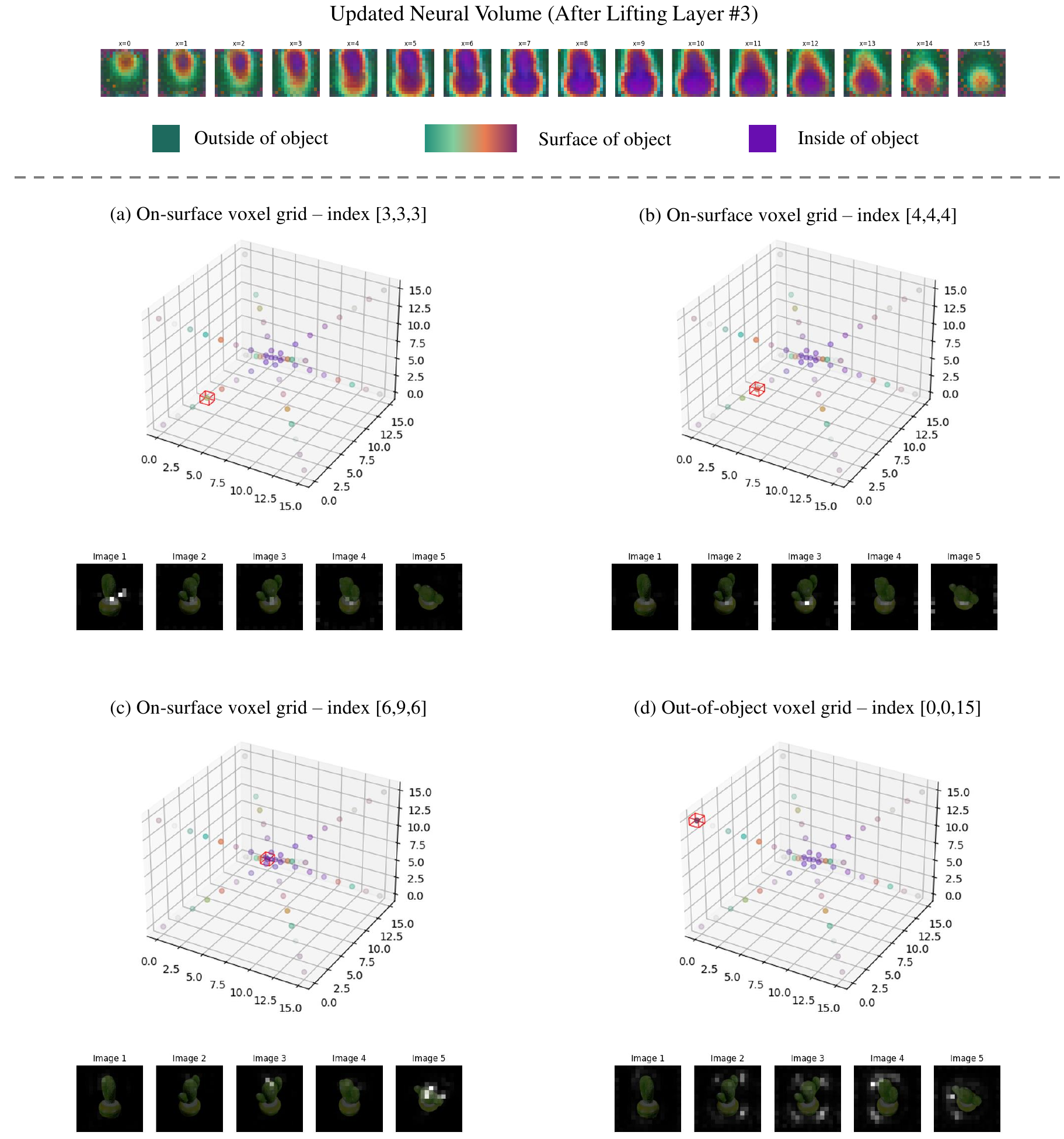}
\caption{\small{\textbf{Visualization of cross-attention weights between the neural volume and the image features in the 2D-to-3D information mapping layer.} (Top) We specify the visualized color for voxel grids outside, on the surface of, and inside the object. (Bottom) Due to the complexity of the 3D volume, we visualize the attention weights for voxel grids on the diagonal of the volume to ensure the traversal of the object surface. The query voxel grids are contained by the red bounding box. (a)-(b) The on-surface voxels attend 2D image features from the same image regions across views. Moreover, the two neighbor voxel grids demonstrate similar attention patterns, showing the smoothness of the learned mapping function. (c) Another example for an on-surface voxel grid. (d) The attention diffuses for a non-surface voxel grid.
}}
%\vspace{-0.2in}
\label{fig: analysis_3d_attn}
\end{figure*}

\paragraph{The 2D-to-3D Information Mapping Module.} We then dive deeper into the 2D-to-3D information mapping module to understand how the neural volume aggregates information from the 2D image features. The results are shown in Fig.~\ref{fig: analysis_3d_attn}.

\section{More Results and Ablations}

\begin{figure}[h]
\centering
\vspace{-0.1in}
\includegraphics[width=\linewidth]{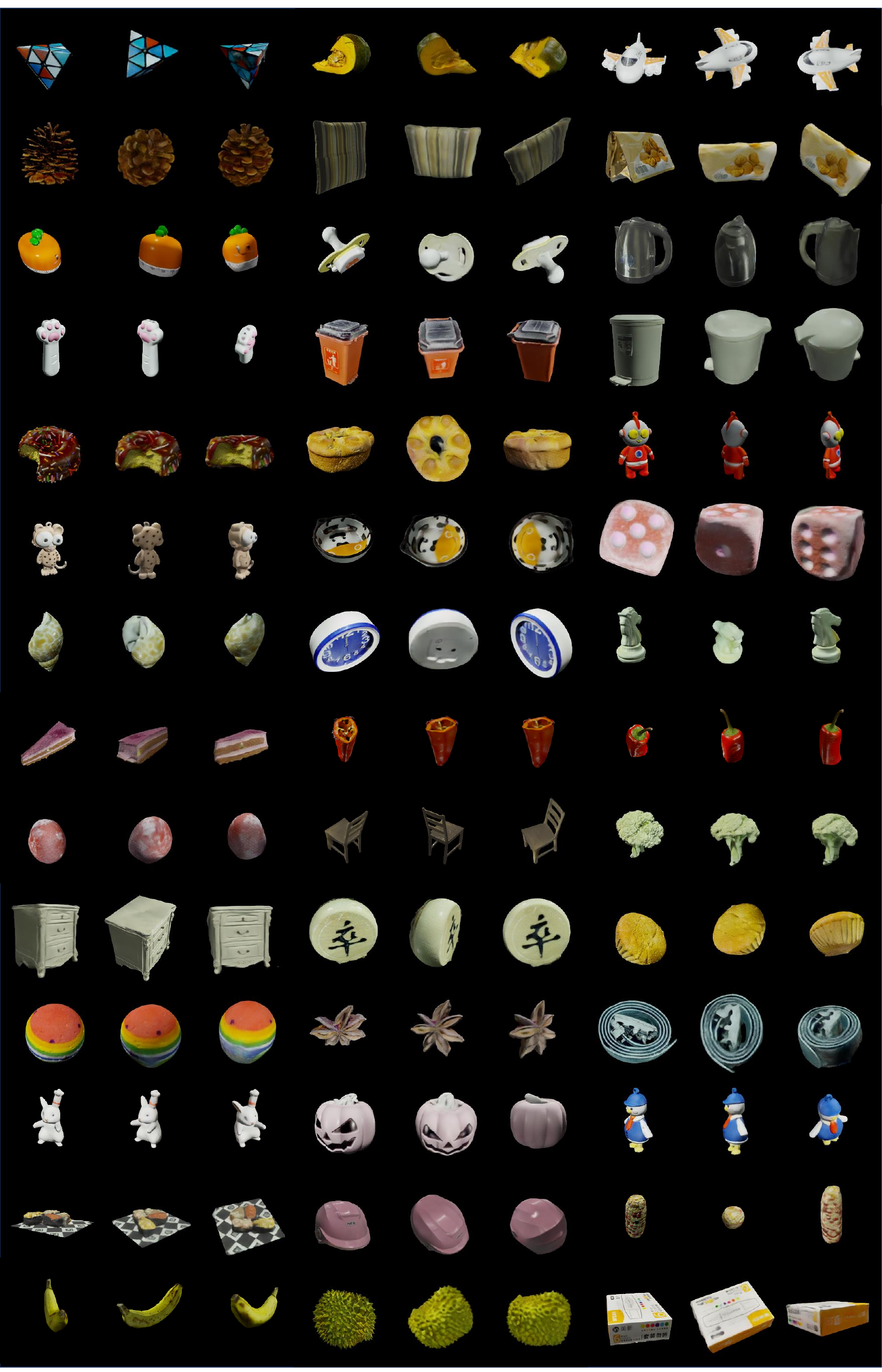}
\vspace{-0.15in}
\caption{\small{\textbf{Visualization of the Omniobject3D dataset.} For each example, we include three images, where the first is one out of five input views, and the last two are rendered novel views.
}}
%\vspace{-0.2in}
\label{fig: vis_supp_omniobject}
\end{figure}

\begin{figure}[h]
\centering
\vspace{-0.2in}
\includegraphics[width=0.9\linewidth]{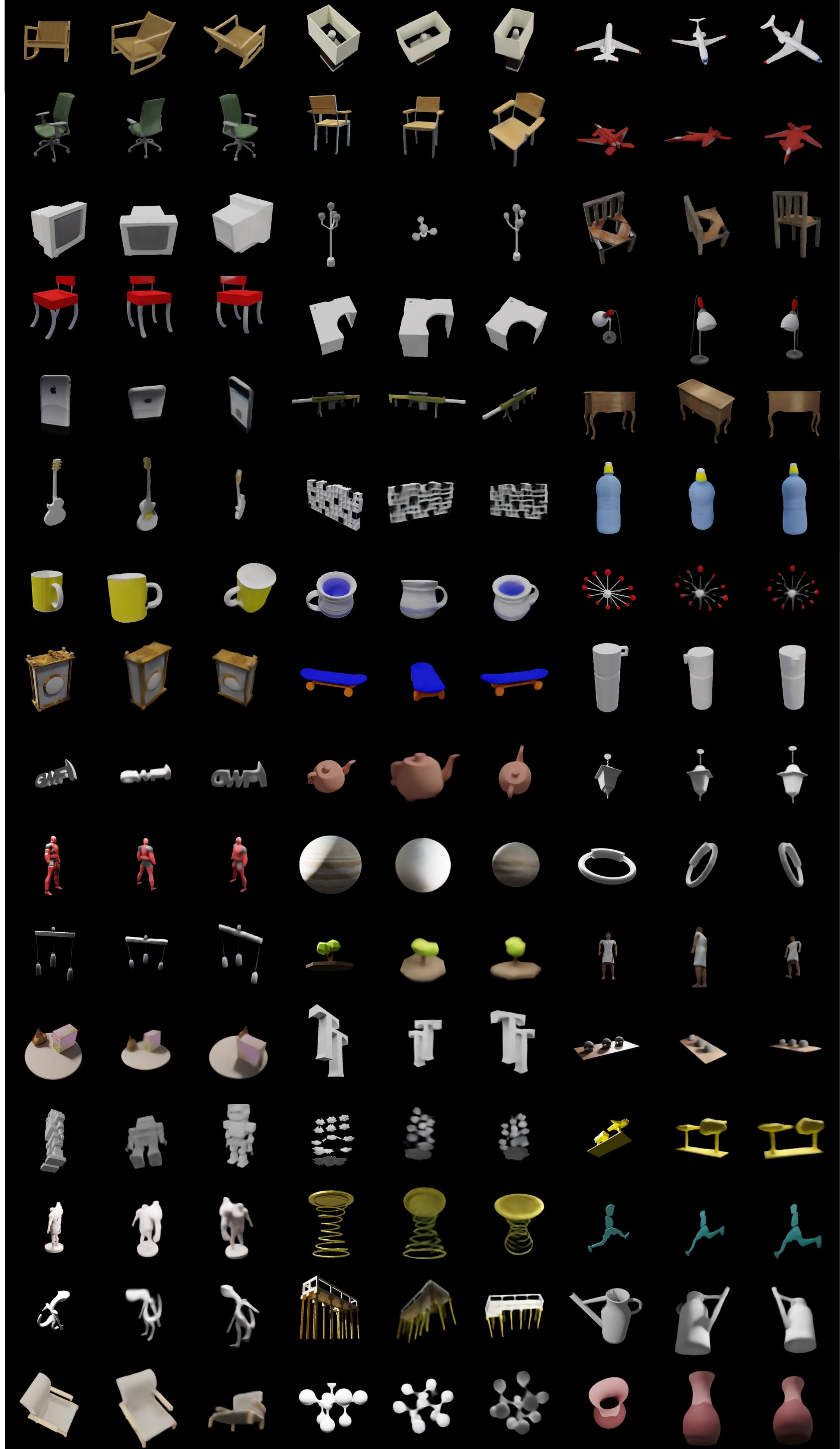}
\vspace{-0.05in}
\caption{\small{\textbf{Visualization of the Kubric (top 8 rows) and Objaverse (bottom 8 rows) dataset.} For each example, we include three images, where the first is one out of five input views, and the last two are rendered novel views.
}}
%\vspace{-0.2in}
\label{fig: vis_supp_kubric_objaverse}
\end{figure}

\begin{figure}[h]
\centering
\vspace{-0.2in}
\includegraphics[width=\linewidth]{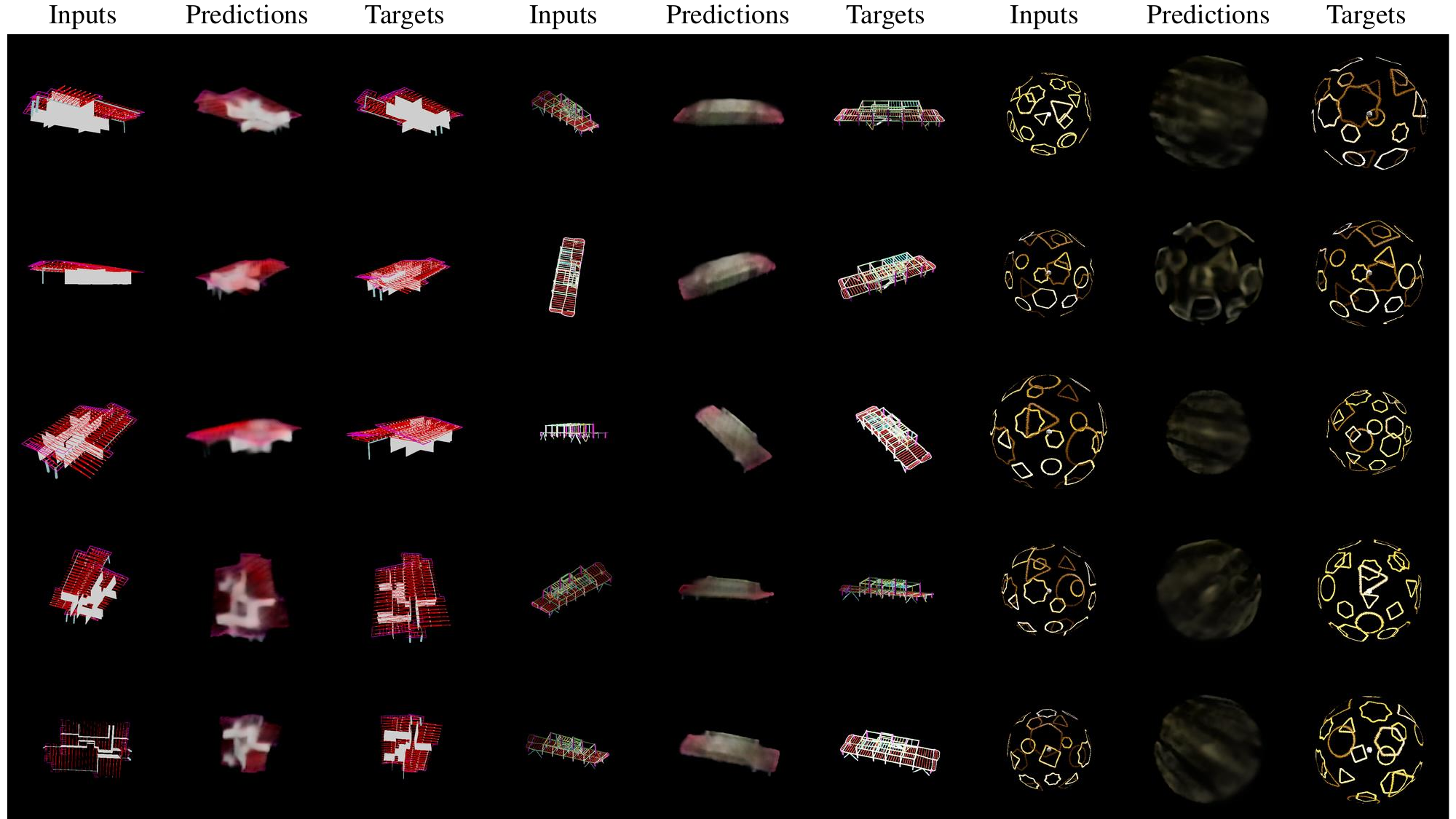}
\vspace{-0.2in}
\caption{\small{\textbf{Visualization of failure cases.} We observe that the performance on objects with very fine-grained geometry details is still limited.
}}
\label{fig: vis_failure}
\end{figure}

We show more visualization of novel view synthesis on the evaluation datasets in Fig.~\ref{fig: vis_supp_omniobject} (Omniobject3D) and Fig.~\ref{fig: vis_supp_kubric_objaverse} (Kurbic and Objaverse). To better understand the performance of \modelname{}, we also include failure cases in Fig.~\ref{fig: vis_failure}.

\begin{figure}[h]
\centering
\includegraphics[width=\linewidth]{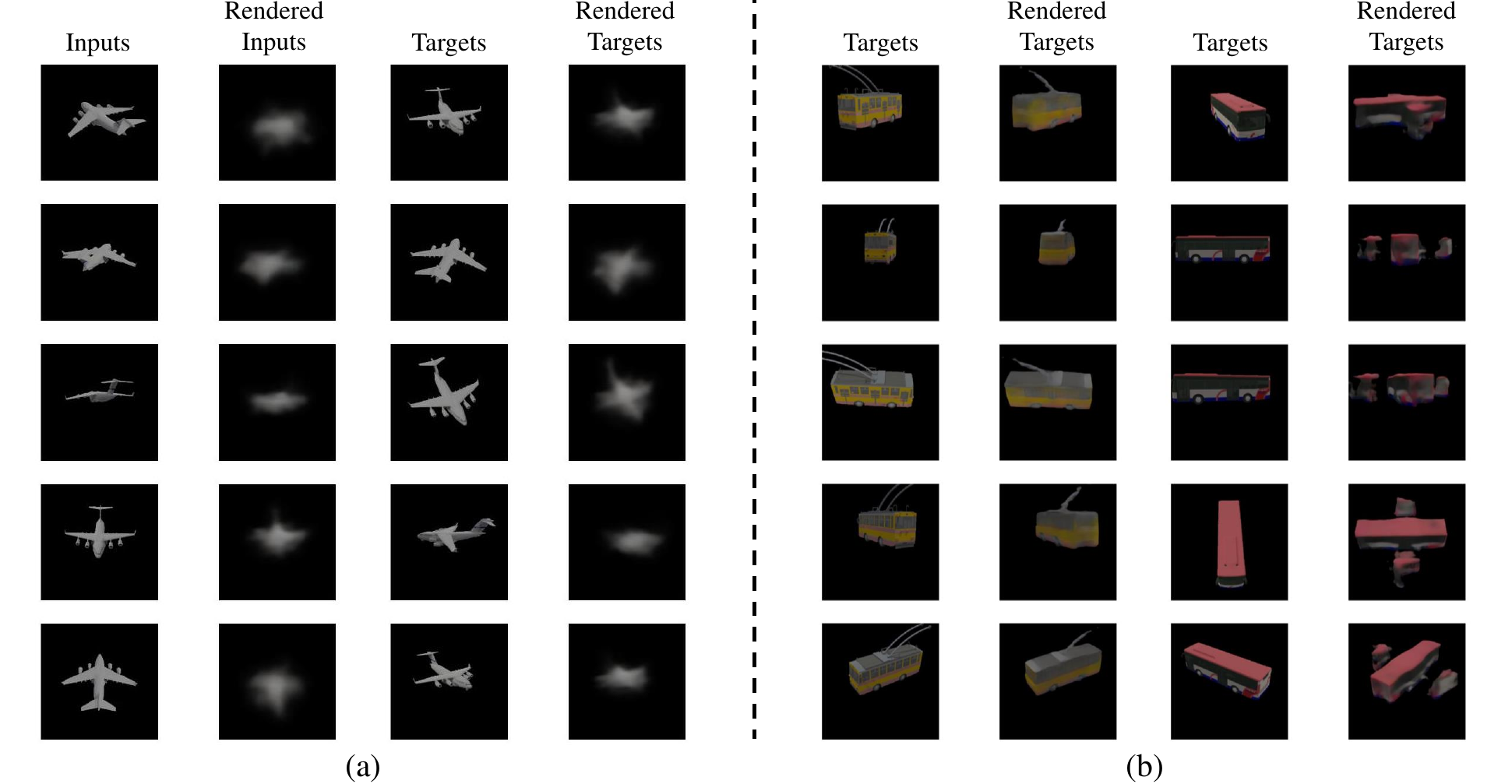}
\vspace{-0.2in}
\caption{\small{\textbf{Visualization of ablation experiments.} (a) Without the multi-view encoder, \modelname{} can only reconstruct noisy results. (b) With using the category-specific coordinate as the world coordinate, \modelname{} degenerates on novel categories. We show two examples, where \modelname{} successfully maps the information in a shared reconstruction space for the first example, but it fails in the second case. 
}}
\label{fig: vis_ablation}
\end{figure}

\begin{figure}[h]
\centering
\includegraphics[width=\linewidth]{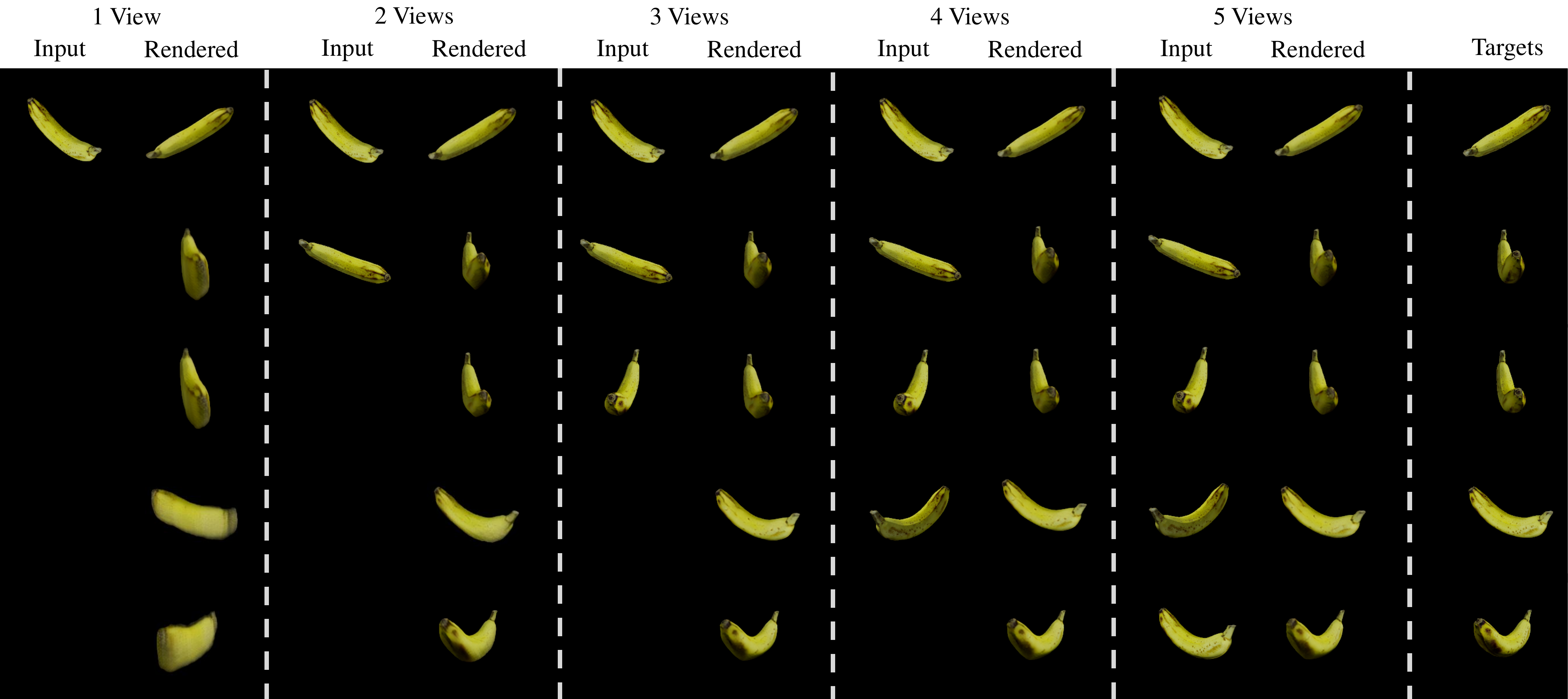}
\vspace{-0.2in}
\caption{\small{\textbf{Visualization of results with different number of input views.} \modelname{} fails to model the object with only one input view, while the performance with two views and five views is close enough. 
}}
\label{fig: vis_num_views}
\end{figure}

Besides, we also include the results for ablation experiments, including the results without using the multi-view encoder (Fig.~\ref{fig: vis_ablation}) and using different numbers of inputs (Fig.~\ref{fig: vis_num_views}). The results verify the significance of the multi-view encoder and the strong capability of \modelname{} for modeling objects from only two views.

\clearpage

\section{Details}
In this section, we introduce the details for training \modelname{} and baselines.

We note that as the neural volume is defined in the local camera coordinate of the canonical view, we render novel views using the relative camera poses rather than the absolute camera poses for training and testing \modelname{}. We keep the same setting for FORGE, SRT, and PixelNeRF. As the definition of world coordinate will not influence the performance of other baseline methods, i.e. SPARF, we use the absolute poses as performed in its official code. Besides, Zero123 does not have the concept of a 3D coordinate system. For FORGE, we perform 5,000-step optimization.

The datasets have different numbers of views for each scene. The object-centric datasets have $100$ (OmniObject3D dataset), $10$ (Kubric dataset), and $12$ views (Objaverse dataset). All views are sampled randomly.
During testing, we use the first $5$ images of each scene (with index 0-4) as inputs, and the other $5$ images (with index 5-9) for evaluation.
For the scene-level DTU dataset, we use images with indexes of 5, 15, 25, 35, and 45 as inputs and use images with indexes of 0, 10, 20, 30, and 40 for testing. This configuration ensures the coverage of the input and evaluation views. For training, we use 5 randomly sampled image sets from each scene as the inputs as well as targets.
We keep the setting for training and testing all baseline methods the same as \modelname{}. During the evaluation, we perform multiple time inference, by setting each input view as canonical. We select the result having the largest PSNR on the input views. Note that the inference speed is calculated for the whole process rather than a single canonical index inference.

\end{document}